\pdfoutput=1

\documentclass[11pt]{article}

\usepackage[preprint]{acl}

\usepackage{times}
\usepackage{latexsym}

\usepackage[T1]{fontenc}

\usepackage[utf8]{inputenc}

\usepackage{microtype}

\usepackage{inconsolata}

\usepackage{graphicx}
\usepackage{bm}
\usepackage{subcaption}
\usepackage{amsmath, amssymb}
\usepackage{array}
\usepackage{xcolor}
\usepackage{algorithm}
\usepackage{algpseudocode}
\usepackage{xcolor}
\usepackage{array, booktabs, makecell}

\definecolor{darkgreen}{RGB}{0, 100, 0}
\definecolor{darkred}{RGB}{139, 0, 0}
\colorlet{myorange}{orange!75!black}

\usepackage{graphicx}
\usepackage{multirow}
\usepackage{float}
\usepackage{amsfonts}
\usepackage{enumitem}
\usepackage{tikz}
\usepackage{tcolorbox}
\usepackage{siunitx}
\usepackage{arydshln}

\usetikzlibrary{calc,fit,positioning,arrows,arrows.meta,backgrounds,decorations.pathreplacing}

\definecolor{c1}{cmyk}{0,0.6175,0.8848,0.1490} 
\definecolor{c2}{cmyk}{0.1127,0.6690,0,0.4431} 
\definecolor{c3}{cmyk}{0.3081,0,0.7209,0.3255} 
\definecolor{c4}{cmyk}{0.6765,0.2017,0,0.0667} 
\definecolor{c5}{cmyk}{0,0.8765,0.7099,0.3647} 

\newtcbox{\hlprimarytab}{on line, rounded corners, box align=base, colback=c3!10,colframe=white,size=fbox,arc=3pt, before upper=\strut, top=-2pt, bottom=-4pt, left=-2pt, right=-2pt, boxrule=0pt}
\newtcbox{\hlsecondarytab}{on line, box align=base, colback=red!10,colframe=white,size=fbox,arc=3pt, before upper=\strut, top=-2pt, bottom=-4pt, left=-2pt, right=-2pt, boxrule=0pt}

\newcommand{\dashifted}{\raisebox{0.5\depth}{\tiny$\downarrow$}}
\newcommand{\uashifted}{\raisebox{0.5\depth}{\tiny$\uparrow$}}
\newcommand{\da}[1]{{\scriptsize\hlprimarytab{\dashifted{#1}}}}
\newcommand{\ua}[1]{{\scriptsize\hlsecondarytab{\uashifted{#1}}}}

\title{Prompt-Based Bias Calibration for Better Zero/Few-Shot Learning of Language Models}

\author{Kang He \quad Yinghan Long \quad Kaushik Roy \\
         Electrical and Computer Engineering, Purdue University \\
         \texttt{\{he603, long273, kaushik\}@purdue.edu}}

\begin{document}
\maketitle
\begin{abstract}
Prompt-based learning is susceptible to intrinsic bias present in pre-trained language models (LMs), leading to sub-optimal performance in prompt-based zero/few-shot settings. In this work, we propose a \textit{null-input prompting} method to calibrate intrinsic bias encoded in pre-trained LMs. Different from prior efforts that address intrinsic bias primarily for social fairness and often involve excessive computational cost, our objective is to explore enhancing LMs' performance in downstream zero/few-shot learning while emphasizing the efficiency of intrinsic bias calibration. Specifically, we leverage a diverse set of auto-selected null-meaning inputs generated from GPT-4 to probe intrinsic bias of pre-trained LMs. Utilizing the bias-reflected probability distribution, we formulate a distribution disparity loss for bias calibration, where we exclusively update bias parameters ($0.1\%$ of total parameters) of LMs towards equal probability distribution. Experimental results show that the calibration promotes an equitable starting point for LMs while preserving language modeling abilities. Across a wide range of datasets, including sentiment analysis and topic classification, our method significantly improves zero/few-shot learning performance of LMs for both in-context learning and prompt-based fine-tuning (on average $9\%$ and $2\%$, respectively).\footnote{Our code is available at \url{https://github.com/kang-ml/prompt_based_bias_calibration}.}
\end{abstract}

\section{Introduction}

The advent of GPT models \cite{radford2019language, brown2020language} has catalyzed the transformative prompt-based learning paradigm. The innovative approach of "pre-train, prompt, and predict" \cite{schick-schutze-2021-exploiting, liu2023pre} facilitates fast adaptation of pre-trained language models (LMs) in learning various tasks and empowers LMs’ strong zero/few-shot learning abilities \cite{schick-schutze-2021-just, gao-etal-2021-making}.

Due to the susceptibility to bias ingrained in pre-trained LMs, prompt-based learning tends to make biased predictions toward some specific answers, 
thereby impacting  performance in prompt-based zero/few-shot settings \cite{ zhao2021calibrate, han2023prototypical}.
To mitigate this issue and improve LM performance, \citet{zhao2021calibrate} and \citet{holtzman2022surface} propose to reweigh LM output probabilities. \citet{han2023prototypical} explores calibrating decision boundaries. While these research has demonstrated substantial improvements, they are primarily designed for in-context learning with frozen pre-trained LMs, leading to two main limitations: (1) They may be not effective in task-specific fine-tuning scenario \cite{jian-etal-2022-contrastive}. Note, however, prompt-based fine-tuning has shown performance improvements over in-context learning \cite{gao-etal-2021-making, logan-iv-etal-2022-cutting}. It is particularly important for relatively small-sized LMs. (2) The intrinsic bias encoded in pre-trained LMs persists since these research focuses on \textit{output calibration} and does not modify LMs.


\begin{figure*}[ht]
\centering
\includegraphics[width=1.0\textwidth]{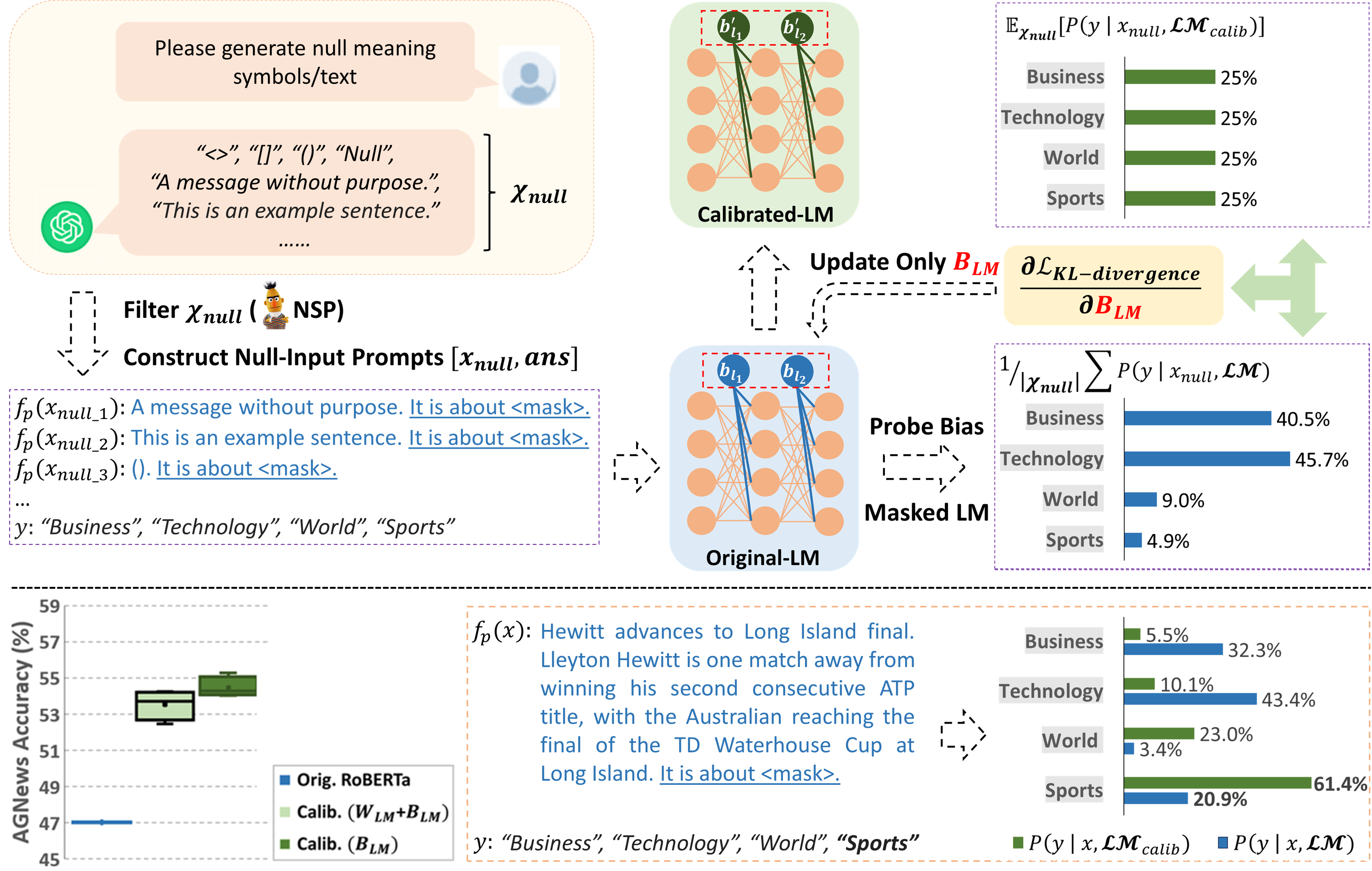}
\caption{We demonstrate our calibration method significantly improves classification performance of pre-trained LM. \textbf{Upper}: The pipeline of proposed null-input prompting method for intrinsic bias calibration targeting AGNews task \cite{zhang2015character}. \textbf{Lower left}: Performance comparison of zero-shot in-context learning using: original LM (Orig. RoBERTa); calibrated (Calib.) LM with full model updates (\textbf{\textit{W\textsubscript{LM}}} + \textbf{\textit{B\textsubscript{LM}}}); calibrated LM with only \textbf{\textit{B\textsubscript{LM}}} updates. \textbf{Lower right}: Case study illustrating that LM makes correct prediction after intrinsic bias calibration.}
\label{fig1}
\end{figure*}

To address these limitations, we investigate the potential for enhancing the performance of LMs as zero/few-shot learners in classification tasks by \textit{calibrating intrinsic bias} of pre-trained LMs. This exploration extends to various prompt-based learning scenarios: in-context learning and prompt-based fine-tuning.
Prior approaches to mitigate intrinsic bias primarily focus on achieving social fairness, and often require laborious corpora augmentation and costly re-training \cite{huang-etal-2020-reducing, kaneko2021debiasing, solaiman2021process, li2023survey}. To improve efficiency in both data generation and model updates, 
we propose leveraging auto-generated \textit{null-meaning inputs} to prompt pre-trained LMs for intrinsic bias probing, and subsequently updating only \textit{bias parameters} \textbf{\textit{B\textsubscript{LM}}} of LMs for bias calibration. Null-meaning inputs are essentially normal text devoid of meaningful content or sentiment. Unlike numerical-zero inputs, they maintain the contextual framework of prompts, ensuring the proper functioning of contextual LMs.
Our motivation stems from the expectation that bias-calibrated models should produce uniform probabilities across all categories if the input in a prompt delivers null information \cite{zhao2021calibrate}. \textbf{\textit{B\textsubscript{LM}}} functions as offsets in neural networks, and strategically updating only \textbf{\textit{B\textsubscript{LM}}} could potentially counteract intrinsic bias of pre-trained models, achieving higher efficiency (updating $\sim0.1\%$ parameters of entire LM). The approach promotes an equitable starting point, and we expect that the light model updates preserve pre-trained models' language modeling abilities while maintaining the focus on bias calibration, ultimately making LMs better zero/few-shot learners.

The pipeline of our calibration method is illustrated in Figure~\ref{fig1}. We use Masked LMs (RoBERTa \citealp{liu2019roberta}) for zero/few-shot learning since they generally produce competitive performance in classification tasks and their moderate size facilitates combining prompting with fine-tuning \cite{gao-etal-2021-making, liu2023pre}. 
First, we utilize GPT-4 API to automatically generate diverse null-meaning inputs $\mathcal{X}_{\text{null}}$ including symbols, words, phrases, and sentences. This generation process is downstream task-agnostic. 
By concatenating each null-meaning input $x_{\text{null}}$ with an answer format $\textit{ans}$ aligned with the downstream task, we construct null-input prompts (similar to \citealp{zhao2021calibrate}), e.g., \textit{"An empty sentence. It is about <mask>."}.
For better cohesive integration of the \textit{"null"} information into the prompts, we additionally devise a filtering strategy to select $x_{\text{null}}$, to which the answer format $\textit{ans}$ exhibits relatively strong Next Sentence Prediction (NSP) correlation \cite{devlin-etal-2019-bert}. Next, we update \textbf{\textit{B\textsubscript{LM}}} with null-input prompts to calibrate intrinsic bias. Given the absence of task-relevant information in these prompts, the anticipated outcome in the parameter updating process is a convergence towards equal output probabilities for each label word. We formulate a customized Kullback–Leibler (KL) divergence loss for gradient descent on \textbf{\textit{B\textsubscript{LM}}} to minimize the distribution disparity. Finally, bias-calibrated LMs are applied in downstream prompt-based zero/few-shot learning following \citet{gao-etal-2021-making}.

The main contributions of our work are:
\begin{itemize}[noitemsep,nolistsep]
  \item We introduce a null-input prompting method for calibrating intrinsic bias of pre-trained Masked LMs, aiming for better prompt-based zero/few-shot classification performance.
  \vspace{3pt}
  \item Our method integrates two key aspects for efficient bias calibration: auto-construction of null-input prompts and updating only bias parameters of LMs.
  The calibration promotes a fair starting point for LMs while preserving language modeling abilities.
  \vspace{3pt}
  \item Extensive experiments on eight classification datasets 
  with four prompt-based learning approaches show that our method significantly improves LMs' zero/few-shot performance, and outperforms output-calibration methods.
\end{itemize}

\section{Related Work}

\label{related work}
\noindent\textbf{Impact of intrinsic bias on downstream LM performance.}
Intrinsic bias in pre-trained LMs stems from imbalances present in extensive pre-training corpora. Higher frequency of specific terms in those corpora could lead to \textit{common token bias} \cite{zhao2021calibrate}. Additionally, frequent co-occurrence of certain terms with specific sentiment in pre-training could introduce \textit{association bias} \cite{cao2022aspect}.
Because of those intrinsic bias, prompt-based predictions by pre-trained LMs are prone to bias towards some specific answers, resulting in sub-optimal performance in downstream tasks \cite{zhao2021calibrate, han2023prototypical}.



\vspace{5pt}
\noindent\textbf{Mitigating strategies.} Research has focused on counteracting the bias solely at the output prediction stage, without modifying pre-trained LMs. For example, \citet{zhao2021calibrate} introduces contextual calibration and \citet{holtzman2022surface} presents Domain Conditional Pointwise Mutual Information to reweigh answer scores. \citet{min-etal-2022-noisy} explores computing the probability of the input conditioned on the label. \citet{han2023prototypical} proposes to calibrate decision boundaries. However, these studies mainly demonstrate their effectiveness for in-context learning using frozen pre-trained LMs, without addressing the intrinsic bias encoded in the LMs. 
Other research on mitigating intrinsic bias primarily targets removing social bias \cite{dinan-etal-2020-queens, huang-etal-2020-reducing, cheng2021fairfil, zhou2023causal}, often employing costly data augmentation and re-training, and as a by-product, degrades language modeling abilities \cite{meade-etal-2022-empirical}.

Efficiently calibrating intrinsic bias in pre-trained LMs for enhancing downstream zero/few-shot learning performance is an open research problem. We introduce a parameter-efficient intrinsic-bias calibration method leveraging automatically constructed null-input prompts, which significantly improves zero/few-shot learning of LMs. 

\vspace{5pt}
\noindent\textbf{Parameter-efficient fine-tuning (PEFT) for downstream tasks.} It has been demonstrated that fine-tuning a very small portion of model parameters can achieve performance on par with fine-tuning the entire set of parameters. People propose integrating small, trainable adapter modules between model layers \cite{bapna-firat-2019-simple, houlsby2019parameter}, coupled with further optimization using low-rank adaptations (LoRA) \cite{hu2021lora}. Some other research focuses on prompt tuning \cite{lester-etal-2021-power, li-liang-2021-prefix, gu-etal-2022-ppt, guo2022improving} which only tunes continuous prompt embeddings for efficiently adapting pre-trained LMs to downstream tasks.

Our method provides a unique perspective of enhancing LM performance on downstream tasks through efficient intrinsic-bias calibration. We update only bias parameters of pre-trained LMs with null-input prompts in calibration. Contrary to adapters and LoRA which would need sufficient labeled data to learn new matrices, we do not introduce new matrices to pre-trained LMs, preserving LMs' few-shot learning capabilities.
Moreover, our approach does not necessarily require target-domain data (whether labeled or unlabeled), enabling fully unsupervised deployment, particularly advantageous for zero-shot setting. 

\section{Null-Input Prompting for Intrinsic Bias Calibration}
\label{calib}
\subsection{Task Formulation}

Let $\bm{\mathcal{LM}}$ be a pre-trained Masked LM. Verbalizer $V(\cdot)$ maps label $y$ to vocabulary token. Prompt function $f_{\textit{p}}(\cdot)$ modifies original input $x_{\text{in}}$ into cloze-style prompt containing one \texttt{<mask>} token to be predicted.
The output representation $\mathbf{h}_\text{<mask>}$ of the \texttt{<mask>} token is acquired from the last encoder layer after forwarding the prompt to the LM. Following \citet{gao-etal-2021-making}, the probability prediction of each class $y \in \mathcal{Y}$ is formulated as:
\begin{multline}
\label{eq_1}
P(y \,|\, x_{\text{in}}, \bm{\mathcal{LM}}) = P(V(y) \,|\, f_{\textit{p}}(x_{\text{in}}), \bm{\mathcal{LM}}) = \\
\frac{\exp\left(\textit{index}_{V(y)}(\mathbf{W}_{\text{lm\_head}} \cdot \mathbf{h}_{\text{<mask>}})\right)}{\sum_{j=1}^{|\mathcal{Y}|} \exp\left(\textit{index}_{V(y_{\text{j}})}(\mathbf{W}_{\text{lm\_head}} \cdot \mathbf{h}_{\text{<mask>}})\right)},
\end{multline}
where $\mathbf{W}_{\text{lm\_head}}$ is the pre-trained \textit{masked language modeling head} weight matrix, and $\textit{index}_{V(y)}$ selects the logits corresponding to the label words based on their index in LM token list.

One can probe intrinsic bias encoded in pre-trained LM by replacing $x_{\text{in}}$ with null-meaning input $x_{\text{null}} \in \mathcal{X}_{\text{null}}$ \cite{zhao2021calibrate}. $\mathcal{X}_{\text{null}}$ represents a set of $x_{\text{null}}$ and we will elaborate their generation and selection in \S~\ref{null}.
As shown by the blue bars in the upper part of Figure~\ref{fig1}, while null-meaning inputs essentially provide no task-relevant prior information, the mean output probability associated with different labels $\bar{P}_{\mathcal{X}_{\text{null}}}(y \,|\, x_{\text{null}}, \bm{\mathcal{LM}})$ may exhibit significant difference attributed to model's intrinsic bias. Ideally, for bias-calibrated LM $\bm{\mathcal{LM}_{\textit{calib}}}$, the expectation of output distribution conditioned on null-meaning inputs should be uniform across all label words, i.e.,

\begin{equation}
\mathbb{E}_{\mathcal{X}_{\text{null}}} \left[ P(y \,|\, x_{\text{null}}, \bm{\mathcal{LM}_{\textit{calib}}}; \forall y \in \mathcal{Y}) \right] =
\frac{1}{\lvert \mathcal{Y} \rvert}.
\label{eq_2}
\end{equation}

We aim to calibrate intrinsic bias by updating LM to minimize this distribution disparity which we quantify using differentiable KL divergence as:
\begin{align}
\label{eq_3}
& D_{\mathcal{KL}}\left(U(\mathcal{Y}) \, || \, \bar{P}_{\mathcal{X}_{\text{null}}}(\mathcal{Y})\right) \nonumber\\
& = \sum_{y \in \mathcal{Y}} \left(1 / \lvert \mathcal{Y} \rvert \cdot \log\frac{1 / \lvert \mathcal{Y} \rvert}{\bar{P}_{\mathcal{X}_{\text{null}}}(y)} \right) \nonumber\\ 
& = \log(1 / \lvert \mathcal{Y} \rvert)-(1 / \lvert \mathcal{Y} \rvert) \cdot \sum_{y \in \mathcal{Y}}\log\bar{P}_{\mathcal{X}_{\text{null}}}(y),
\end{align}
where $U(\mathcal{Y})$ denotes uniform probability distribution and $\bar{P}_{\mathcal{X}_{\text{null}}}(y)$ represents the simplified form of $\bar{P}_{\mathcal{X}_{\text{null}}}(y \,|\, x_{\text{null}}, \bm{\mathcal{LM}})$. 

\subsection{Update Only Bias Parameters}
While intrinsic bias may be encoded across various parts of pre-trained LMs, one question arises: is it essential to update the entire model, or is there a more efficient alternative that can achieve comparable effectiveness in intrinsic bias calibration? We propose to only update bias parameters \textbf{\textit{B\textsubscript{LM}}}, with the following rationale: (\romannumeral 1) \textbf{\textit{B\textsubscript{LM}}} constitutes less than $0.1\%$ of total LM parameters, offering significant memory and computation cost saving compared to updating entire LM. (\romannumeral 2) Weight parameters \textbf{\textit{W\textsubscript{LM}}}\footnote{\textbf{\textit{W\textsubscript{LM}}} also includes embedding parameters in our context.} may carry crucial pre-existing knowledge for language modeling, which risks impairment with a full model update \cite{meade-etal-2022-empirical}. \textbf{\textit{B\textsubscript{LM}}}, often overlooked in LM research, serves as offsets in neural network layers. Strategic updates may counteract intrinsic bias while potentially preserving language modeling abilities. (\romannumeral 3) Empirical research on efficient fine-tuning has demonstrated the important role of bias parameters in LMs \cite{ben-zaken-etal-2022-bitfit, logan-iv-etal-2022-cutting}.

We update \textbf{\textit{B\textsubscript{LM}}} using gradient descent to minimize the dissimilarity between output probability distribution from the LM conditioned on null-meaning inputs and uniform probability distribution $U(\mathcal{Y})$. We formulate a customized KL divergence loss $\mathcal{L}$, including both divergence of individual null-input's output distribution $P_i(\mathcal{Y})$ with respect to $U(\mathcal{Y})$, and batch-averaged distribution $\bar{P}_{N}(\mathcal{Y})$ with respect to $U(\mathcal{Y})$, as:
\begin{align}
\label{eq_5}
\mathcal{L} = & \frac{1}{N} \sum_{i=1}^{N} D_{\mathcal{KL}}\bigl(U(\mathcal{Y}) \,||\, P_i(\mathcal{Y})\bigr) \nonumber\\ 
              & + D_{\mathcal{KL}}\bigl(U(\mathcal{Y}) \,||\, \bar{P}_{N}(\mathcal{Y})\bigr),
\end{align}
where $N$ is the batch size of null-meaning inputs. Incorporating the second term in the loss function promotes calibration stability and aligns with the objective of Equation~\ref{eq_2}.

\subsection{Early Stopping of Calibration}
\label{early stop}

We aim to obtain LM with improved zero/few-shot performance at the calibration stopping point. An overly calibrated model may simply produce uniform probability predictions regardless of input information. 
To avoid this, we develop specialized early stopping strategies depending on whether the downstream task is zero-shot or few-shot.

\vspace{5pt}
\noindent\textbf{For \textit{zero-shot} downstream tasks.}
Determining the calibration stopping point for optimal zero-shot learning performance is challenging due to the absence of labeled data for validation during calibration. To discern the patterns of a good stopping point, we first conduct empirical experiments by validating LM zero-shot performance on the entire test dataset after each calibration batch (consisting of $N$ null-meaning inputs) across different calibration learning rates (Figure~\ref{fig7} in Appendix~\ref{details}).
As shown in Figure~\ref{fig2}, with optimal calibration learning rate, model performance exhibits significant improvements in the first one/few calibration batches with low variance, and then starts to degrade and becomes unstable. The low performance and instability at the calibration tail confirm our assumption on the detrimental effects of excessive calibration on LM's modeling abilities. Notably, calibration with only one batch of null inputs (indicated by the red vertical line in Figure~\ref{fig2}) delivers consistent and significant improvement compared to the original LM (although might not be the best improvement). Therefore, for enhancing LM zero-shot performance, we directly adopt the \textit{One-batch Calibration} as the early stopping criterion. 

\begin{figure}[ht]
  \centering
  \includegraphics[width=1\linewidth]{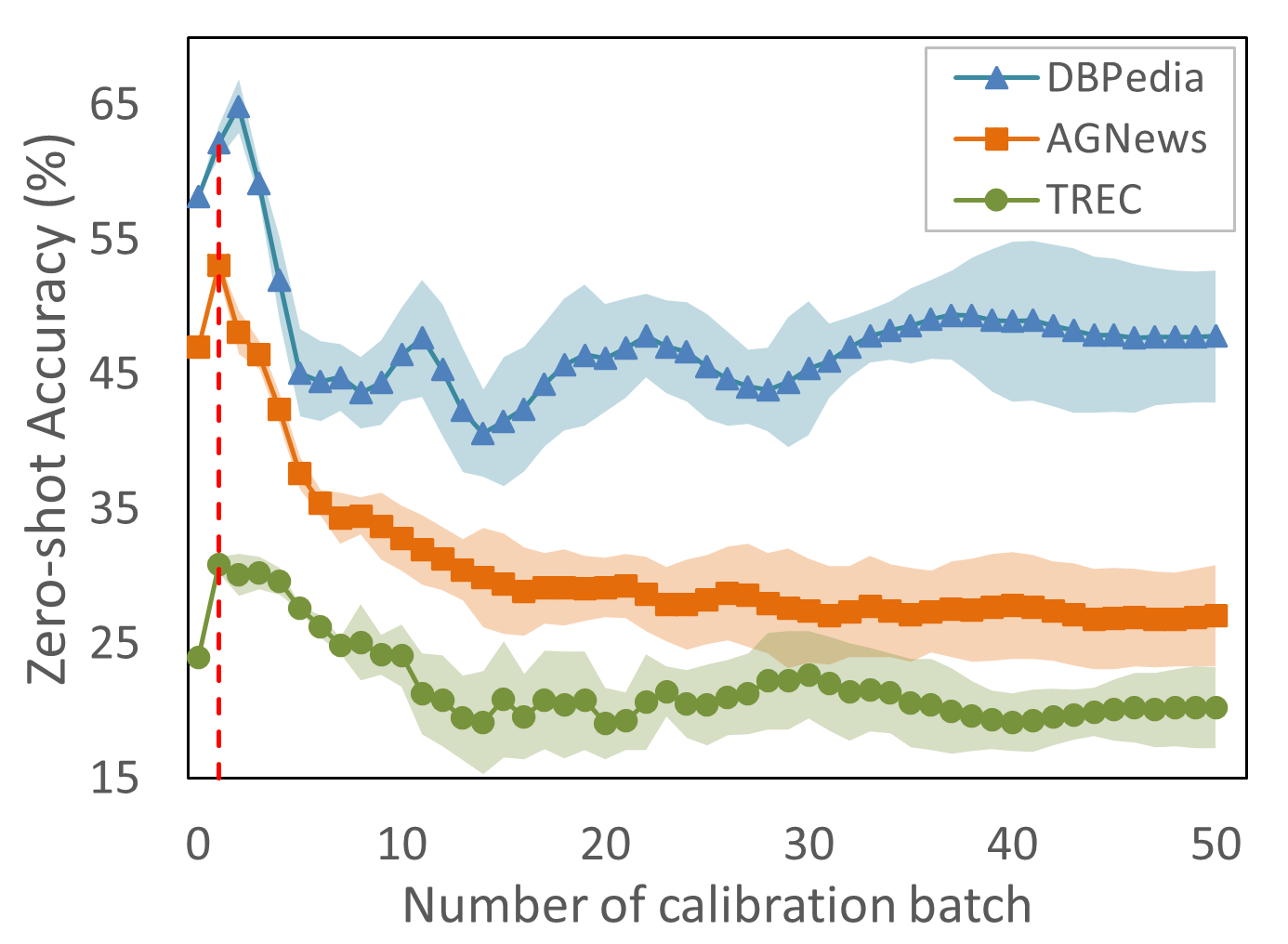}
  \caption{Empirical experiments show the impact of calibration on zero-shot learning performance as the number of calibration batches increases (batch size is 32). 
  The intersections of the curves and red vertical line signify the outcomes of the first calibration batch.}
  \label{fig2}
\end{figure}


\noindent\textbf{For \textit{few-shot} downstream tasks.}
With the acquisition of a few labeled downstream data, the previous challenge of lacking validation for determining the stopping point in the calibration process is alleviated. We utilize the small amount of labeled data as validation dataset $\mathcal{D}_\text{val}^{\text{calib}}$ to set a stopping criterion for calibration. Additionally, we take into account above-mentioned empirical findings that, for some tasks, stopping at one batch of calibration yields optimal LM performance. Relying on the limited size of $\mathcal{D}_\text{val}^{\text{calib}}$ might fail to identify such stopping points. To this effect, we store both $LM_{\text{calib}}^{\text{one\_batch}}$ (obtained from one-batch stopping) and $LM_{\text{calib}}^{\text{val}}$ (obtained from validation-based stopping) for downstream few-shot leaning tasks. Since $LM_{\text{calib}}^{\text{one\_batch}}$ is stored in the process of obtaining $LM_{\text{calib}}^{\text{val}}$, this will not result in additional computation overhead. Memory overhead is minimal, as it only requires storing an additional set of updated bias parameters.

We summarize our method for intrinsic bias calibration in Algorithm~\ref{alg1} (Appendix~\ref{details}). 

\section{Auto-Construct Null-Input Prompt}
\label{null}

\subsection{Generate Null-Meaning Input} 
We employ null-meaning inputs to probe the intrinsic bias of pre-trained LMs, and then use those bias-reflected outputs to calibrate the LMs. Crafting a diverse set of null-meaning inputs $\mathcal{X}_{\text{null}}$ for an averaged output helps prevent overfitting to sub-optimal instances, thereby contributing to the effectiveness of calibration.
To enable cost-effective acquisition of various null-meaning data, we utilize GPT-4 API for automatic generation with instructions such as \textit{"Please generate null meaning symbols, words, phrases, and sentences, in total <Number>."}. This process is task-agnostic, generating data that contains null information with respect to any downstream task. Note that null information is not equivalent to neutral sentiment, as it carries no inherent meaning or contextual sentiment implications. We further validate this through t-SNE \cite{JMLR:v9:vandermaaten08a} visualization in Appendix~\ref{details} Figure~\ref{fig6}.

\begin{table}[ht]
  \centering
  \small
  \begin{tabular}{cc}
    \toprule
    \textbf{Generated null-meaning input} $x_{\text{null}}$ & $P_\textit{nsp}(x_{\text{null}}, \textit{ans})$ \\
    \midrule
    \textit{This is an example sentence.} & \textcolor{darkgreen}{0.9996} \\[0.5ex]
    \textit{A message without purpose.} & \textcolor{darkgreen}{0.9979} \\[0.5ex]
    \textit{Words without message.} & \textcolor{darkgreen}{0.9809} \\[0.001ex]
    \midrule
    \textit{123abc} & \textcolor{myorange}{0.0267} \\[0.5ex]
    \textit{\!@\#\$\%\^{}\&*()-\_=+[]\{\}} & \textcolor{myorange}{0.0145} \\[0.5ex]
    \textit{////////////////////} & \textcolor{myorange}{0.0008} \\[0.001ex]
    \bottomrule
  \end{tabular}
  \caption{Some examples of generated null-mean inputs. In this case, \textit{"It is about <mask>."} is used as the answer format \textit{ans}. The green/yellow numbers represent high/low NSP probabilities, respectively.}
  \label{table1}
\end{table}

\subsection{Select \texorpdfstring{\bm{$x_{\text{null}}$}}{x\_null} and Build Null-Input Prompt} 
\label{null_prompt}
We construct null-input prompt $f_{\textit{p}}(x_{\text{null}})$
by concatenating the generated null-meaning input with an answer format $\textit{ans}$. For consistency, the answer format (e.g., \textit{"It is <mask>."}) is the same as the one intended for use in the downstream task. Some examples are shown in the upper part of Figure~\ref{fig1}. 

\begin{table*}[t]
    \centering
    \small
    \setlength{\tabcolsep}{3pt}
    \begin{tabular}{c cccc cccc cccc ccc}
        \toprule
        \multirow{2.5}{*}{\textbf{}} & 
        \multicolumn{3}{c}{\textbf{In-context lrn no demo\textsuperscript{\dag}}} & & 
        \multicolumn{3}{c}{\textbf{In-context lrn with demo}} & &
        \multicolumn{3}{c}{\textbf{Prompt FT no demo}} & &
        \multicolumn{3}{c}{\textbf{Prompt FT with demo}} \\
        \\[-2ex]
        \cline{2-4} \cline{6-8} \cline{10-12} \cline{14-16}
        \\[-1.5ex]
        & NoCal & OutCal & IntrCal & & 
        NoCal & OutCal & IntrCal & &
        NoCal & OutCal & IntrCal & &
        NoCal & OutCal & IntrCal \\
        \midrule
        AGNews & $47.0_\text{0.0}$ & $54.3_\text{1.0}$ & $\textbf{54.5}_\text{0.6}$ & & $79.7_\text{0.8}$ & $78.8_\text{3.3}$ & $\textbf{82.4}_\text{0.9}$ 
        & & $\textbf{89.1}_\text{0.9}$ & $86.3_\text{1.6}$ & $89.0_\text{0.8}$
        & & $86.9_\text{2.8}$ & $87.5_\text{1.3}$ & $\textbf{89.3}_\text{0.9}$ \\[0.5ex]
        
        DBPedia & $58.2_\text{0.0}$ & $54.1_\text{1.9}$ & $\textbf{61.8}_\text{0.6}$ 
        & & $92.6_\text{0.6}$ & $94.0_\text{0.9}$ & $\textbf{94.8}_\text{0.7}$
        & & $98.2_\text{1.3}$ & $99.0_\text{0.5}$ & $\textbf{99.0}_\text{0.1}$ 
        & & $98.6_\text{0.3}$ & $98.5_\text{0.2}$ & $\textbf{98.9}_\text{0.3}$ \\[0.5ex]
        
        TREC & $24.0_\text{0.0}$ & $29.4_\text{2.1}$ & $\textbf{31.1}_\text{0.5}$ 
        & & $48.3_\text{1.4}$ & $42.5_\text{3.4}$ & $\textbf{48.6}_\text{2.2}$
        & & $85.0_\text{7.4}$ & $82.2_\text{2.0}$ & $\textbf{89.3}_\text{4.5}$
        & & $87.6_\text{2.5}$ & $74.2_\text{4.0}$ & $\textbf{89.7}_\text{1.0}$ \\[0.5ex]
        
        Subj & $50.8_\text{0.0}$ & $\textbf{64.0}_\text{2.7}$ & $62.7_\text{0.8}$ 
        & & $47.2_\text{0.2}$ & $55.0_\text{1.3}$ & $\textbf{63.5}_\text{2.3}$ 
        & & $91.2_\text{0.9}$ & $88.2_\text{2.5}$ & $\textbf{93.2}_\text{1.2}$ 
        & & $91.4_\text{3.3}$ & $93.0_\text{0.8}$ & $\textbf{94.3}_\text{0.2}$ \\[0.5ex]
        
        SST-5 & $31.5_\text{0.0}$ & $33.0_\text{2.1}$ & $\textbf{37.5}_\text{0.4}$ 
        & & $34.4_\text{1.7}$ & $31.2_\text{2.6}$ & $\textbf{36.6}_\text{1.0}$ 
        & & $47.8_\text{4.6}$ & $45.3_\text{2.8}$ & $\textbf{49.9}_\text{2.7}$ 
        & & $47.1_\text{1.9}$ & $42.6_\text{4.0}$ & $\textbf{50.0}_\text{1.7}$ \\[0.5ex]
        
        Laptop & $54.6_\text{0.0}$ & $58.3_\text{2.5}$ & $\textbf{59.6}_\text{1.9}$  
        & & $50.8_\text{1.0}$ & $65.1_\text{2.7}$ & $\textbf{67.4}_\text{1.7}$ 
        & & $74.3_\text{1.4}$ & $74.3_\text{1.6}$ & $\textbf{74.9}_\text{2.9}$ 
        & & $76.8_\text{1.0}$ & $75.6_\text{1.4}$ & $\textbf{78.7}_\text{1.4}$ \\[0.5ex]
        
        Restaurant & $68.6_\text{0.0}$ & $72.0_\text{4.9}$ & $\textbf{72.8}_\text{1.6}$ 
        & & $69.8_\text{1.1}$ & $\textbf{74.3}_\text{1.6}$ & $74.0_\text{1.0}$
        & & $79.7_\text{2.2}$ & $79.0_\text{1.0}$ & $\textbf{82.0}_\text{0.9}$ 
        & & $78.4_\text{4.9}$ & $79.0_\text{5.5}$ & $\textbf{79.8}_\text{4.5}$ \\[0.5ex]
        
        Twitter & $19.7_\text{0.0}$ & $43.4_\text{4.1}$ & $\textbf{51.7}_\text{0.4}$  
        & & $21.0_\text{0.5}$ & $40.7_\text{5.4}$ & $\textbf{49.4}_\text{2.7}$ 
        & & $51.7_\text{2.9}$ & $44.1_\text{3.9}$ & $\textbf{57.0}_\text{4.2}$  
        & & $57.7_\text{2.8}$ & $50.3_\text{4.2}$ & $\textbf{59.3}_\text{2.3}$ \\[0.001ex]
        
        \midrule
        Average & $44.3$ & $51.1$ & $\textbf{54.0}$
        & & $55.5$ & $60.2$ & $\textbf{64.6}$ 
        & & $77.1$ & $74.8$ & $\textbf{79.3}$ 
        & & $78.1$ & $75.1$ & $\textbf{80.0}$ \\
        \bottomrule
    \end{tabular}
    \caption{Result comparisons among NoCal (LM-BFF \citealp{gao-etal-2021-making}; no calibration), OutCal (output calibration) and IntrCal (ours; intrinsic-bias calibrated LM) using RoBERTa-large. We report the mean and standard deviation of performance in 8 classification datasets with 4 prompt-based learning methods. "In-context lrn" refers to in-context learning and "Prompt FT" refers to prompt-based fine-tuning. "with/no demo" denotes incorporating/not incorporating demonstrations in prompts. In-context lrn no demo\textsuperscript{\dag} is zero-shot learning, while the other three are few-shot learning.}
    \label{table2}
\end{table*}

To pursue better cohesive integration of the \textit{"null"} information into the prompts, we prioritize the null-meaning inputs, with which the answer format exhibits higher Next Sentence Prediction (NSP) probability \cite{devlin-etal-2019-bert}. Specifically, after we generate a large set of null-meaning inputs $\{x_{\text{null\_1}}, x_{\text{null\_2}}, \ldots, x_{\text{null\_k}}\}$ and the answer format $\textit{ans}$ is selected, we employ BERT-large model \cite{devlin-etal-2019-bert} to predict NSP $P_\textit{nsp}(x_{\text{null}}, \textit{ans})$ and sort null-meaning inputs by their probabilities. Table~\ref{table1} shows some generated $x_{\text{null}}$, with which a specific answer format presents high/low NSP scores. After the sorting, we retain the top $80\%$ $x_{\text{null}}$ instances (800 in total), which maintains the diversity among the selected samples. We observe that null inputs with low NSP scores are typically randomly-combined alphabet letters and symbols. These samples may have minimal occurrences in pre-training corpora. The low NSP scores can be attributed to RoBERTa’s lack of comprehension of their meanings in context. Their representations extracted by LM might have high variance, which might impact the stability and effectiveness of calibration. We show calibration with the $x_{\text{null}}$ selection strategy further improves LM performance in \S~\ref{results} Table~\ref{table4}.

\section{Experiments}

\label{dataset}
 We conduct extensive experiments on 8 English datasets, including sentiment analysis and topic classification.\footnote{ We mainly focus on single-sentence tasks, which aligns with the use of single-sentence null inputs for calibration. The alignment may enhance calibration effectiveness. We also experiment on sentence-pair tasks in Appendix~\ref{other_exp} Table~\ref{table17} and demonstrate better performance after calibration.}
 They consist of 5 sentence-level datasets potentially impacted by \textit{common token bias}: AGNews \cite{zhang2015character}, DBPedia \cite{lehmann2015dbpedia}, TREC \cite{voorhees2000building}, Subj \cite{pang2004sentimental}, SST-5 \cite{socher2013recursive} and 3 aspect-level sentiment analysis datasets likely subject to \textit{association bias}: Restaurant and Laptop reviews from SemEval 2014 Task \cite{pontiki-etal-2014-semeval}, Twitter \cite{dong-etal-2014-adaptive}. 
 For aspect-level datasets, the task is to predict sentiments associated with the marked aspects in each sentence. More details are in Appendix~\ref{details} Table~\ref{table7}.
 
\subsection{Evaluation Protocol}
We evaluate the effectiveness of our intrinsic-bias calibration method on enhancing Masked LMs zero/few-shot learning performance with 4 prompt-based learning methods: in-context learning and prompt-based fine-tuning, both with and without demonstration. We follow the prompt-based fine-tuning and demonstration method of \citet{gao-etal-2021-making}. Besides Masked LMs, we also validate the effectiveness of our method on two decoder LMs: GPT-2 XL (1.5B) \cite{radford2019language} and Llama-2 (7B) \cite{touvron2023llama} in Appendix~\ref{decoder}.

We conduct calibration with 5 different seeds, and for the few-shot setting, we randomly sample 5 different groups of training and validation sets ($K$ samples per class). We report the mean and standard deviation of LM performance. For the 5 sentence-level classification tasks, we use \textit{accuracy} as the metric. For the 3 aspect-level classification tasks, because of the imbalance in test set, we use \textit{weighted F$_1$} for a balanced evaluation. Details of calibration and prompt-based learning are in Appendix~\ref{details}.

We present our main results using RoBERTa-large, and $K=16$ for few-shot setting. 
Results of using RoBERTa-base, $K=\{2, 4, 8\}$, and different prompt templates are in Appendix~\ref{other_exp} (Table~\ref{table8}, Table~\ref{table9} and Figure~\ref{fig8}).

\subsection{Main Results}
\label{results}

In Table~\ref{table2}, we compare our results of \textbf{IntrCal} (intrinsic bias calibration) with reproduced results of:

\noindent (1) \textbf{NoCal}: No calibration. Use LM-BFF \cite{gao-etal-2021-making} to compute $P(y \,|\, x_{\text{in}})$ for predictions.

\noindent (2) \textbf{OutCal}: Output calibration. OutCal computes $\frac{P(y \,|\, x_{\text{in}})}{P(y \,|\, x_{\text{domain}})}$ instead of $P(y \,|\, x_{\text{in}})$ to counteract surface form competition and bias \cite{zhao2021calibrate, holtzman2022surface}. Note that OutCal was originally demonstrated for in-context learning with GPT models, while here, we apply the method in Masked LMs for fair comparisons.

In addition to NoCal and OutCal, we compare our results with those reproduced from \textit{NoisyTune} \cite{wu-etal-2022-noisytune}, \textit{NSP-BERT} \cite{sun-etal-2022-nsp} and \textit{Perplection} \cite{lu-etal-2023-makes}, as detailed in Appendix~\ref{compare} (Table~\ref{table16}, ~\ref{table5}). The superior performance further validates the effectiveness of our method.

\vspace{5pt}
\noindent\textbf{In-context learning results.} OutCal has significantly improved LM zero/few-shot performance compared to NoCal. Our method (IntrCal) further outperforms OutCal by a large margin: $2.9\%$ and $8.3\%$ absolute in zero-shot learning \& $4.4\%$ and $8.7\%$ absolute in few-shot learning, in terms of average and best-case improvement. This demonstrates the advantages of intrinsic bias calibration over attempting to counteract bias solely at the output. Moreover, OutCal exhibits higher variance in performance due to its sensitivity to human-crafted domain-relevant strings $x_{\text{domain}}$. Using certain $x_{\text{domain}}$ instances may not accurately capture the bias of LMs, resulting in under-calibration or over-calibration and leading to the high variance. In our approach, we use a large set of auto-generated and selected $x_{\text{null}}$ as the training set for bias calibration. This mitigates the impact of sub-optimal samples and enhances calibration robustness, contributing to more stable and reliable performance.

\vspace{5pt}
\noindent\textbf{Prompt-based fine-tuning results.}
This method fine-tunes all LM parameters utilizing limited labeled data by minimizing the cross-entropy loss based on Equation~\ref{eq_1}.
It greatly raises LM performance compared to in-context learning and sets up a strong baseline (i.e., NoCal). OutCal fails to surpass NoCal. We speculate that OutCal's limitation lies in its exclusive focus on offsetting bias at the output and lack of interaction with the interior of LM. This appears to impede OutCal from adapting effectively to the intricate dynamics of LM after prompt-based fine-tuning, leading to some counterproductive calibrations. In contrast, IntrCal (ours) with the aim of intrinsic bias calibration achieves superior performance with absolute gains of maximum $5.3\%$ and average $2\%$ compared to NoCal.

\begin{table}[t]
    \centering
    \small
    \setlength{\tabcolsep}{2pt}
    \begin{tabular}{c cc c cc}
        \toprule
        \multirow{2.5}{*}{\textbf{}} & 
        \multicolumn{2}{c}{\textbf{In-context lrn no demo}} & &
        \multicolumn{2}{c}{\textbf{Prompt FT no demo}} \\
        \\[-2ex]
        \cline{2-3} \cline{5-6}
        \\[-1.5ex]
        & \textit{UnSel.} $x_{\text{null}}$ & \textit{Sel.} $x_{\text{null}}$ & & 
        \textit{UnSel.} $x_{\text{null}}$ & \textit{Sel.} $x_{\text{null}}$\\
        \midrule
        AGNews & $53.1_\text{0.6}$ & $\textbf{54.5}_\text{0.6}$ 
        & & $87.8_\text{1.7}$ & $\textbf{89.0}_\text{0.8}$  \\[0.5ex]
        
        DBPedia & $\textbf{62.1}_\text{1.2}$ & $61.8_\text{0.6}$ 
        & & $98.7_\text{0.2}$ & $\textbf{99.0}_\text{0.1}$\\[0.5ex]
        
        TREC & $30.9_\text{0.6}$ & $\textbf{31.1}_\text{0.5}$ 
        & & $88.5_\text{3.5}$ & $\textbf{89.3}_\text{4.5}$\\[0.5ex]
        
        Subj & $60.5_\text{3.2}$ & $\textbf{62.7}_\text{0.8}$ 
        & & $92.8_\text{1.6}$ & $\textbf{93.2}_\text{1.2}$\\[0.5ex]
        
        SST-5 & $35.5_\text{1.7}$ & $\textbf{37.5}_\text{0.4}$ 
        & & $48.7_\text{4.2}$ & $\textbf{49.9}_\text{2.7}$\\

        \bottomrule
    \end{tabular}
    \caption{Calibration with selected null-meaning inputs ($x_{\text{null}}$) further improves LM performance. \textit{UnSel.} refers to using $x_{\text{null}}$ without selection, while \textit{Sel.} denotes using selected $x_{\text{null}}$ based on the sorting of $P_\textit{nsp}(x_\text{null}, \textit{ans})$ (\S~\ref{null_prompt}).}
    \label{table4}
\end{table}

\begin{figure}[ht]
  \centering
  \includegraphics[width=1\linewidth]{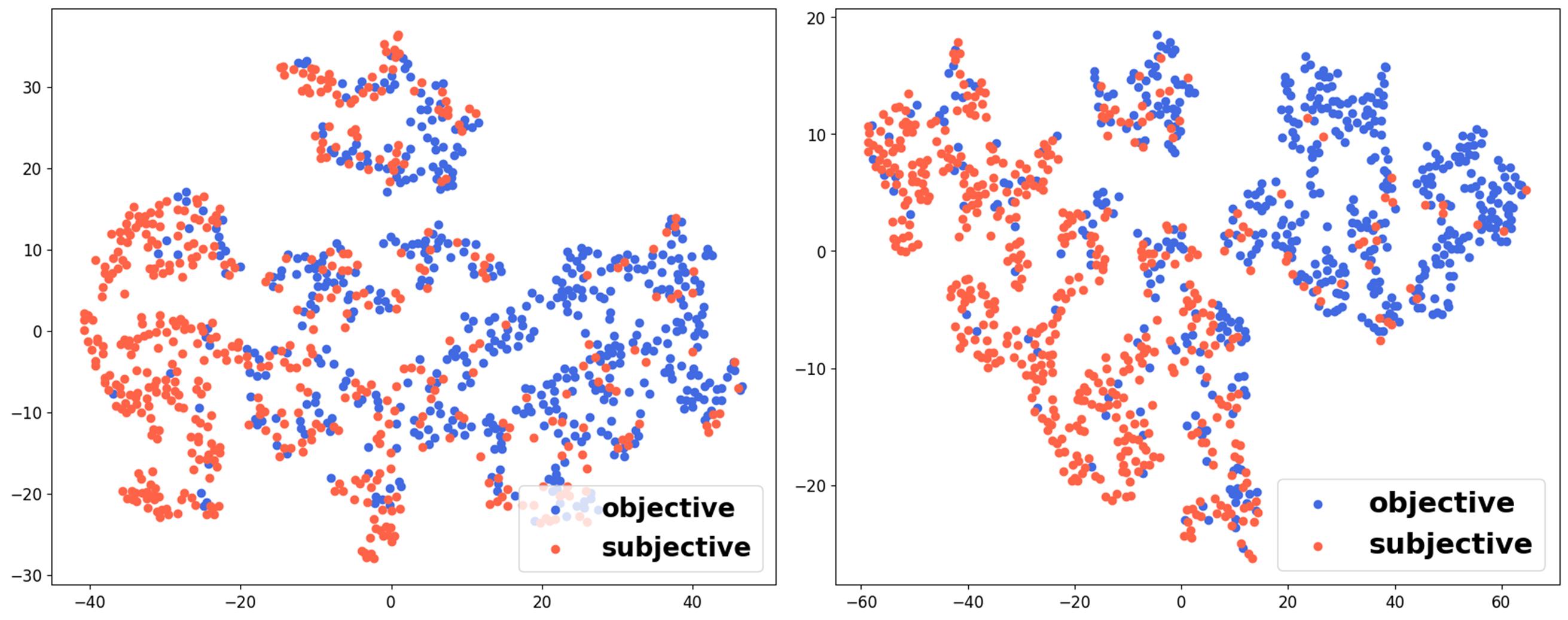}
  \caption{t-SNE visualization for output representations of \texttt{<mask>} token. \textbf{Left} is obtained from original LM; \textbf{Right} is obtained from the LM after \textit{One-batch Calibration}. Two colors denote the two classes in Subj task.}
  \label{fig4}
\end{figure}

The output representations of \texttt{<mask>} token for label word predictions are visualized by t-SNE in Figure~\ref{fig4}. On the left, samples from the two categories are almost mixed together, indicating that the original LM tends to bias toward one class prediction. In contrast, the right visualization demonstrates improved separability after \textit{One-batch Calibration}(\S~\ref{early stop}), which explains the significant performance enhancement achieved by our intrinsic-bias calibration method.


\subsection{Update Entire LM vs. Only Bias Parameters in Calibration}
\label{update}

In Table~\ref{table3}, we evaluate the impact of updating entire LM (\textbf{\textit{W\textsubscript{LM}}} + \textbf{\textit{B\textsubscript{LM}}}) during calibration
on downstream task performance, as compared to only updating bias parameters (\textbf{\textit{B\textsubscript{LM}}}). 
The optimal learning rate for updating entire LM is smaller (Appendix~\ref{details} Table~\ref{table6}). For in-context learning, the LM with only \textbf{\textit{B\textsubscript{LM}}} updates in calibration achieves better overall performance compared to the LM with entire parameter updates, most likely attributed to better preserved language modeling abilities (Appendix~\ref{other_exp} Table~\ref{table14}).
For prompt-based fine-tuning, two differently calibrated LMs demonstrate comparable performance, as the impact of entire-parameter calibration on the modeling ability is mitigated through task-specific fine-tuning. Considering the significant saving in memory and computation, we recommend only updating \textbf{\textit{B\textsubscript{LM}}} in calibration.

\begin{table}[H]
    \centering
    \small
    \setlength{\tabcolsep}{2pt}
    \begin{tabular}{c cc c cc}
        \toprule
        \multirow{2.5}{*}{\textbf{}} & 
        \multicolumn{2}{c}{\textbf{In-context lrn no demo}} & &
        \multicolumn{2}{c}{\textbf{Prompt FT no demo}} \\
        \\[-2ex]
        \cline{2-3} \cline{5-6}
        \\[-1.5ex]
        & \textbf{\textit{W\textsubscript{LM}}} + \textbf{\textit{B\textsubscript{LM}}} & \textbf{\textit{B\textsubscript{LM}}} & & 
        \textbf{\textit{W\textsubscript{LM}}} + \textbf{\textit{B\textsubscript{LM}}} & \textbf{\textit{B\textsubscript{LM}}}\\
        \midrule
        AGNews & $53.5_\text{0.8}$ & $\textbf{54.5}_\text{0.6}$ 
        & & $\textbf{89.3}_\text{0.8}$ & $89.0_\text{0.8}$\\[0.5ex]
        
        DBPedia & $\textbf{63.2}_\text{0.9}$ & $61.8_\text{0.6}$
        & & $99.0_\text{0.5}$ & $\textbf{99.0}_\text{0.1}$\\[0.5ex]
        
        TREC & $\textbf{31.3}_\text{0.8}$ & $31.1_\text{0.5}$
        & & $87.6_\text{2.8}$ & $\textbf{89.3}_\text{4.5}$\\[0.5ex]
        
        Subj & $53.3_\text{0.6}$ & $\textbf{62.7}_\text{0.8}$
        & & $\textbf{93.7}_\text{0.6}$ & $93.2_\text{1.2}$\\[0.5ex]
        
        SST-5 & $33.5_\text{0.4}$ & $\textbf{37.5}_\text{0.4}$ 
        & & $49.4_\text{0.7}$ & $\textbf{49.9}_\text{2.7}$\\[0.5ex]
        
        Laptop & $58.2_\text{0.8}$ & $\textbf{59.6}_\text{1.9}$ 
        & & $\textbf{78.1}_\text{1.3}$ & $74.9_\text{2.9}$\\[0.5ex]
        
        Restaurant & $70.7_\text{1.8}$ & $\textbf{72.8}_\text{1.6}$ 
        & & $81.3_\text{1.0}$ & $\textbf{82.0}_\text{0.9}$\\[0.5ex]
        
        Twitter & $\textbf{51.8}_\text{0.7}$ & $51.7_\text{0.4}$ 
        & & $55.7_\text{2.3}$ & $\textbf{57.0}_\text{4.2}$\\
        
        \midrule
        Average & $51.9$ & $\textbf{54.0}$
        & & $79.3$ & $79.3$ \\
        \bottomrule
    \end{tabular}
    \caption{Performance comparisons between differently calibrated LMs. \textbf{\textit{W\textsubscript{LM}}} + \textbf{\textit{B\textsubscript{LM}}} updates entire LM in calibration while \textbf{\textit{B\textsubscript{LM}}} only updates bias parameters. Additional results of In-context lrn/Prompt FT \textit{with demo} are in Appendix~\ref{other_exp} Table~\ref{table11}.}
    \label{table3}
\end{table}

\subsection{Analysis}
\begin{figure}[ht]
  \centering
  \includegraphics[width=1\linewidth]{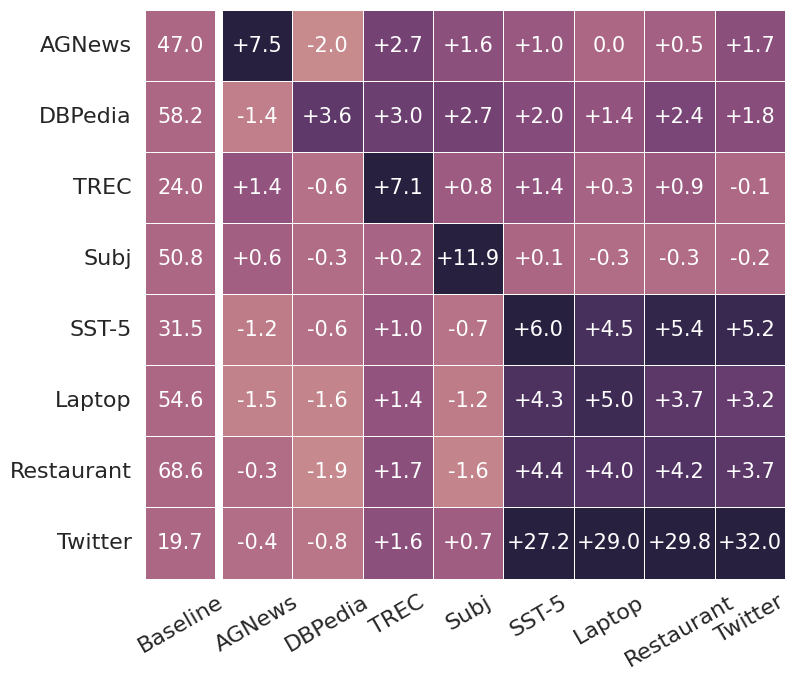}
  \caption{Impact of calibration on downstream tasks shown through the changes with respect to baseline on each column. Each row shows the zero-shot performance of one task employing: \textit{original LM} (first column; baseline), \textit{task-specific calibrated LM} (diagonal), \textit{other-task calibrated LM} (other places).}
  \label{fig5}
\end{figure}

\noindent\textbf{How does intrinsic bias calibration impact downstream tasks?}
Our method calibrates the intrinsic bias associated with a set of task-specific label words. 
In this section, we explore the impact of updating LM for task-specific bias calibration on other downstream task performance.
Specifically, we take the LM calibrated for one task and evaluate its performance on the other tasks as shown in Figure~\ref{fig5}.
In general, intrinsic bias calibration for one task has a minimal adverse effect on other tasks' performance (no more than 2\% degradation) because of the light model updates, while remarkably enhancing LM performance on that specific task. Notably, there is consistent performance increase at bottom right, as these tasks are all sentiment classification sharing or including same label words.\footnote{For aspect-level datasets, better improvement is on the diagonals (task-specific calibration), indicating our method mitigates the impact of association bias (Appendix~\ref{details}).}

\vspace{5pt}
\noindent\textbf{How does intrinsic bias calibration impact language modeling abilities?} 
We employ pseudo-perplexity \cite{salazar-etal-2020-masked} to evaluate language modeling for Masked LM. Following each task-specific intrinsic bias calibration, we measure pseudo-perplexity and compare the results with original RoBERTa on WikiText-2, WikiText-103 \cite{merity2017pointer}, and LAMBADA dataset \cite{paperno-etal-2016-lambada}. As shown in Table~\ref{table13}, language modeling abilities are largely preserved after calibration due to the minimal updates to the model. 

\begin{table}[H]
\centering             
\small
\setlength{\tabcolsep}{2.1pt}
\begin{tabular}{lrrr}
\toprule



\textbf{} & \textbf{WT-2} & \textbf{WT-103} & \textbf{LAMBADA}\\
\midrule
Original RoBERTa                           & $6.189$ &            $7.008$ &           $24.52$ \\[0.5ex]
\, + \textsc{calibration} & & \\[0.5ex]
\, \, \, \text{for\_AGNews} &     \ua{0.017} $6.206$ & \ua{0.029} $7.037$ & \ua{0.02} $24.54$ \\[0.5ex]
\, \, \, \text{for\_DBPedia} &    \ua{0.008} $6.197$ & \ua{0.002} $7.010$ & \da{0.22} $24.30$ \\[0.5ex]
\, \, \, \text{for\_TREC} &       \da{0.027} $6.162$ & \da{0.042} $6.966$ & \da{0.27} $24.25$ \\[0.5ex]
\, \, \, \text{for\_Subj} &       \da{0.021} $6.168$ & \da{0.030} $6.978$ & \ua{0.08} $24.60$ \\[0.5ex]
\, \, \, \text{for\_SST-5} &      \da{0.031} $6.158$ & \da{0.039} $6.969$ & \da{0.18} $24.34$ \\[0.5ex]
\, \, \, \text{for\_Laptop} &     \ua{0.011} $6.200$ & \ua{0.002} $7.010$ & \da{0.01} $24.51$ \\[0.5ex]
\, \, \, \text{for\_Restaurant} & \ua{0.055} $6.244$ & \ua{0.074} $7.082$ & \ua{0.13} $24.65$ \\[0.5ex]
\, \, \, \text{for\_Twitter} &    \da{0.029} $6.160$ & \da{0.037} $6.971$ & \ua{0.05} $24.57$ \\[0.0001ex]  
\bottomrule
\end{tabular}
\caption{Pseudo-perplexities of \textit{original RoBERTa} and \textit{task-specific calibrated RoBERTa} on WikiText-2 (WT-2), WikiText-103 (WT-103) and LAMBADA. We use 2000 test samples of each dataset. An increase in values (highlighted in red) indicates a reduction in language modeling abilities after calibration.}
\label{table13}
\end{table}

\section{Conclusion}
In this work, we propose a null-input prompting method to calibrate the intrinsic bias of pre-trained Masked LMs, aiming to enhance zero/few-shot learning performance in classification tasks. Our method incorporates two key features for efficiency: (1) auto-construction of null-input prompts for bias probing, leveraging a diverse set of selected null-meaning inputs easily crafted from generative Large LM; (2) updating only bias parameters for bias calibration. 
Experimental results show that bias-calibrated LMs demonstrate significant performance improvement for both in-context learning and prompt-based fine-tuning, with average gains of $9\%$ and $2\%$, respectively.
Moreover, our method outperforms output-calibration approaches, highlighting the advantage of intrinsic bias calibration. 
We believe this work presents a new perspective of making LMs better zero/few-shot learners via intrinsic bias calibration.  Additionally, the demonstrated significance of bias parameters could provide insights for future bias-related research.

\section*{Limitations}
While our method has achieved substantial improvement in prompt-based zero/few-shot learning, it comes with limitations that could open avenues for future research. 

First, calibration is fully unsupervised in the scenario where no labeled data is available (zero-shot downstream tasks in \S~\ref{early stop}). Based on empirical experimental results, we adopt the conservative \textit{One-batch Calibration} strategy to ensure a safe and consistent performance enhancement.
In the future, we aim to explore more rigorous approaches to determine optimal stopping points in this scenario.

Second, we utilize RoBERTa (encoder) models for classification tasks, as encoder models may more effectively encode task-specific patterns for discriminative tasks
compared to some generative LMs \cite{gao-etal-2021-making, li2023label}, as shown in Table~\ref{table15}. However, the relatively small size of those Masked LMs (355M parameters for RoBERTa-large) could be the ultimate limitation to their capabilities.
Given the proliferation of large-scale generative (decoder) LMs and their accomplishments in tackling more challenging tasks \cite{thoppilan2022lamda, chowdhery2023palm, touvron2023llama}, we anticipate extending our method to large decoder models and validating the applicability of our findings. Furthermore, we expect to expand the scope of tasks to include regression problems (e.g., sentiment score prediction) leveraging KL divergence to measure disparities in continuous probability distributions, aiming to address bias-related challenges across diverse scenarios.

\section*{Ethics Statement and Broader Impact}
Our work is conformant to the Code of Ethics. We appropriately cite relevant methods, models, and datasets that we use. We affirm that all datasets in our experiments are public, and no private or sensitive information is incorporated in our research. Our use of datasets and pre-trained models is consistent with their intended use. For broader impacts, our method, extending beyond calibrating common token bias and association bias, might inspire prospective research in mitigating social bias and improving the fairness of pre-trained LMs.

\section*{Acknowledgments}
This work was supported in part by the Center for Co-Design of Cognitive Systems (CoCoSys), a Semiconductor Research Corporation (SRC) and DARPA-sponsored JUMP 2.0 center.

\bibliography{anthology, custom}
\clearpage

\appendix
\section{Experimental Details}
\label{details}

\textbf{Prompts with or without demonstrations.} Table~\ref{table7} shows the prompt templates and label words of each dataset we use for main experiments. 

For downstream tasks, in few-shot setting, task-specific example-label pairs (i.e., demonstrations) can be incorporated in the context to enhance the LM's comprehension. While in zero-shot setting, no labeled data is available and thereby no demonstrations. 

For calibration, demonstrations are either absent from or added to null-input prompts, consistent with their exclusion from or inclusion in prompts for downstream tasks.  An example of a null-input prompt without demonstration is:

\vspace{4pt}
\begin{tabular}{l}
  \hspace{-0.8em} 
  \textbf{<s> \textit{An empty sentence.}} \hspace{0.6em} 
  \textbf{\textit{\underline{It is <mask>.} </s>}} \\
\end{tabular}

\vspace{4pt}
\noindent \textit{<s>} and \textit{</s>} respectively denote \texttt{<cls>} token and \texttt{<sep>} token in RoBERTa. In the other case, we incorporate demonstrations retrieved from the small training set into the null-input prompt such as:

\vspace{4pt}
\begin{tabular}{l}
  \hspace{-0.8em} 
  \textbf{<s> \textit{An empty sentence.}} \hspace{0.6em} 
  \textbf{\textit{\underline{It is <mask>.} </s>}} \\[0.3ex]
  
  \hspace{-0.8em}
  \textit{Compellingly watchable.} \hspace{0.3em} 
  \textit{\underline{It is great.} </s>} \\[0.3ex]

  \hspace{-0.8em}
  \textit{The film is strictly routine.}  \textit{\underline{It is terrible.} </s>} \\
\end{tabular}

\vspace{10pt}
\noindent \textbf{Association-bias calibration for aspect-level task.} 
For aspect-level sentiment analysis, e.g., \textit{"Wonderful food but poor \underline{service}. \underline{Service} was <mask>."}, the answer contains the aspect word \textit{"service"}. Because the model makes sentiment predictions for specific aspect words, the task is likely subject to \textit{association bias} (\S~\ref{related work}). For association-bias calibration, the only difference is that we incorporate various aspect words in the answer format (e.g., \textit{"<aspect words> was <mask>."}) when constructing null-input prompts. One can either leverage GPT-4 to generate in-domain aspect words (e.g., for restaurant reviews, the generated aspect words could be \textit{menu, food}, etc.), or simply employ the aspect words in the original training dataset. In this work, we choose the latter option. Due to the variability of \textit{<aspect words>} in the answer format, sorting null-meaning inputs by NSP score can yield different results. To this effect, we do not apply $x_{\text{null}}$ selection strategy (\S~\ref{null_prompt}) for aspect-level task, and instead keep all the generated $x_{\text{null}}$.

\vspace{10pt}
\noindent \textbf{Null-meaning inputs generation with GPT-4.} The version of GPT-4 used in our experiment is \texttt{gpt-4-0613}. We observe that GPT-4 could generate repetitive null-meaning inputs. To avoid overrepresentation of certain null inputs which might impact the diversity and introduce bias to the null-input set, we adopt an iterative approach. In each iteration, GPT-4 generates 500 null-meaning inputs, and duplicates are removed. This process continues until we obtain 1000 distinct null-meaning inputs, which takes 3 iterations in our experiment.

\begin{figure}[ht]
  \centering
  \includegraphics[width=1\linewidth]{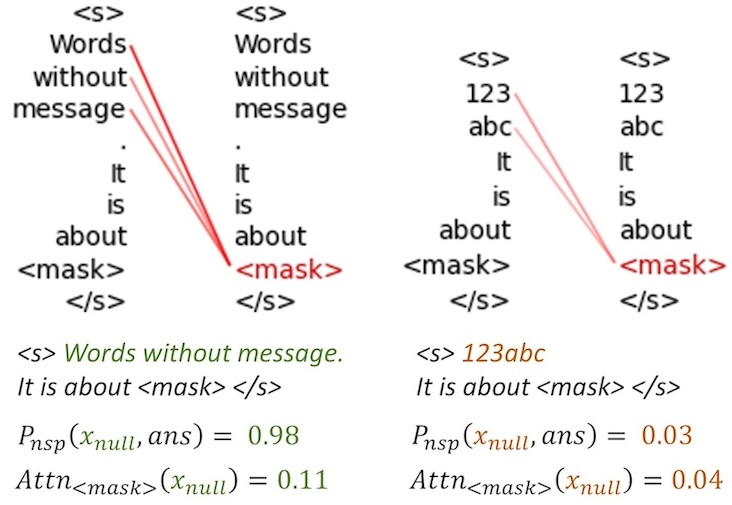}
  \caption{Visualization of attention score by the depth of color in the connecting lines. We only show the attention between \texttt{<mask>} token and null-meaning input $x_{\text{null}}$. $Attn_{\textit{<mask>}}(x_{\text{null}})$ is the attention score of \texttt{<mask>} on $x_{\text{null}}$, averaged over encoder layers and attention heads. \textbf{Left}: Higher attention score indicates enhanced pattern extraction from $x_{\text{null}}$ which has higher $P_\textit{nsp}(x_{\text{null}}, \textit{ans})$.}
  \label{fig3}
\end{figure}

\vspace{10pt}
\noindent \textbf{Null-meaning inputs for \textit{One-batch Calibration}.} For zero-shot downstream tasks, since only one batch of null-meaning inputs is required for calibration in our early-stopping criterion (\S~\ref{early stop}), we select the $Top\text{-}N\{P_\textit{nsp}(x_{\text{null}}, \textit{ans})\}$ $x_{\text{null}}$ from $\mathcal{X}_{\text{null}}$, where $N$ is batch size. We prioritize these samples as our observations show that null-meaning inputs with higher $P_\textit{nsp}(x_{\text{null}}, \textit{ans})$ exhibit higher attention scores between the null input and \texttt{<mask>}, as demonstrated in Figure~\ref{fig3}. 
This indicates more effective conveyance of the \textit{"null"} information to the placeholder \texttt{<mask>}, which could facilitate LM deciphering the \textit{"null"} patterns of the prompts and benefit calibration.

\vspace{10pt}
\noindent \textbf{Hyper-parameters.} In calibration stage, we shuffle the null-input prompts and conduct gradient descent on \textbf{\textit{B\textsubscript{LM}}} (or \textbf{\textit{W\textsubscript{LM}}} + \textbf{\textit{B\textsubscript{LM}}} as comparative experiment) with 5 different seeds to account for calibration variance. There are two main hyper-parameters for calibration: (1) $x_\text{null}$ batch size $N$; (2) calibration learning rate $lr_{calib}$. We conduct grid search on $N=\{8, 16, 32\}$ and $lr_{calib}$ from $1e-6$ to $5e-3$, and obtain the best settings: $N=32$ and $lr_{calib}$ as shown in Table~\ref{table6}.

Calibrated LMs are applied in downstream tasks with prompt-based learning methods. We use the same hyper-parameters as \citet{gao-etal-2021-making} for prompt-based learning. We evaluate on each task’s original test set, except for AGNews and  DBPedia, where we randomly sample 800 test examples.

We use PyTorch \cite{paszke2019pytorch} and public HuggingFace Transformers library \cite{wolf-etal-2020-transformers}. RoBERTa related experiments are conducted on a single NVIDIA V100 GPU, while GPT-2 and Llama-2 experiments are conducted on one A100 GPU in Google Colab.

\begin{table}[H]
\centering
\small
\begin{tabular}{c c c c}
    \toprule
    \multicolumn{1}{c}{\textbf{}} & \multicolumn{2}{c}{\textbf{Calibration ($lr_{calib}$)}} & \multicolumn{1}{c}{\multirow{2.5}{*}{\textbf{\thead{Prompt FT \\ (downstream)}}}} \\ 
    \\[-2ex]
    \cline{2-3}
    \\[-1.5ex]
    \multicolumn{1}{c}{\textbf{}} & \textbf{\textit{W\textsubscript{LM}}} + \textbf{\textit{B\textsubscript{LM}}} & \textbf{\textit{B\textsubscript{LM}}}  & \multicolumn{1}{c}{}\\ 
    \midrule
    
    No demo & $1e-5$ & $1e-3$ & $1e-5$ \\[1ex]
    With demo & $1e-6$ & $1e-4$ & $1e-5$ \\[0.001ex]
    
    \bottomrule
\end{tabular}
\caption{Optimal learning rates for calibration and downstream prompt-based fine-tuning (Prompt FT). With/No demo denotes adding/not adding demonstrations in prompts.}
\label{table6}
\end{table}

\begin{table*}[t]
    \centering
    \small
    \setlength{\tabcolsep}{2pt}
    \begin{tabular}{lccc}
        \toprule
        \textbf{Dataset} & \textbf{Task Type} & \textbf{Prompt Template} & \textbf{Label Words} \\
        \midrule
        AGNews & News topic classification & \{Sentence\} It is about \texttt{<mask>}. & World / Sports / Business / Technology\\[1ex]
        
        DBPedia\textsuperscript{\dag} & Ontology classification & \{Sentence\} It is about \texttt{<mask>}. & Company / Artist / Building / Nature\\[1ex]
        
        TREC & Question classification & \{Sentence\} It is about \texttt{<mask>}. & \thead{Number / Location / Person \\ / Description / Entity / Expression}\\[1ex]
        
        Subj & Subjectivity classification & \{Sentence\} This is \texttt{<mask>}. & objective / subjective\\[1ex]
        
        SST-5 & Movie sentiment analysis & \{Sentence\} The movie was \texttt{<mask>}. & terrible / bad / okay / good / great\\[1ex]
        
        Laptop & Aspect level sentiment analysis & \{Sentence\} \{Aspect words\} was \texttt{<mask>}. & terrible / okay / great\\[1ex]
        
        Restaurant & Aspect level sentiment analysis & \{Sentence\} \{Aspect words\} was \texttt{<mask>}. & terrible / okay / great\\[1ex]
        
        Twitter & Aspect level sentiment analysis & \{Sentence\} \{Aspect words\} was \texttt{<mask>}. & terrible / okay / great\\
        
        \bottomrule
    \end{tabular}
    \caption{Prompt templates and label words of the eight datasets in our experiments for main results. For DBPedia\textsuperscript{\dag}, we use four classes out of the total fourteen classes.}
    \label{table7}
\end{table*}

\begin{algorithm}[ht]
\caption{Null-input prompting for calibration}
\label{alg1}
\begin{algorithmic}[1]
    \Statex \textbf{Inputs:}
    \Statex Downstream task: \textit{zero\_shot} or \textit{few\_shot}
    \Statex Null-input prompts: $\{N_\text{prompt}\}$ 
    \Statex (Val. data in Calibration: $\mathcal{D}_\text{val}^{\text{calib}} \leftarrow \mathcal{D}_\text{train}^{\text{downstrm}})$ 
    \Comment{Only when downstream task is \textit{few\_shot}.} \,\,\,
    \Comment{Downstream training dataset $\mathcal{D}_\text{train}^{\text{downstrm}}$ constitutes $K$ samples per class.}
    
    \Statex \textbf{Output:}
    \Statex $LM_{\text{calib}}^{\text{one\_batch}}$ for \textit{zero\_shot}
    \Statex $LM_{\text{calib}}^{\text{one\_batch}}$ \& $LM_{\text{calib}}^{\text{val}}$ for \textit{few\_shot}

    \For {$batch$ in $\{N_{\text{prompt}}\}$}
        \State $P$ = $\mathcal{LM}$($batch$) \Comment{Null input prompting} 
        \State $\mathcal{L}$ = $D_{\mathcal{KL}}(U \,||\, P)$ \Comment{Unif. distribution $U$}
        \State $\textbf{\textit{B\textsubscript{LM}}} \leftarrow \textbf{\textit{B\textsubscript{LM}}} - lr_{calib} \cdot \frac{\partial \mathcal{L}}{\partial \textbf{\textit{B\textsubscript{LM}}}}$ 
        \If{$first \ batch$}
            \State Save $LM_{\text{calib}}^{\text{one\_batch}}$
        \EndIf
        \If{downstream is \textit{zero\_shot}}
            \textbf{break}
        \EndIf
        \If{better $Compute\_Metric$($\mathcal{D}_\text{val}^{\text{calib}}$)}
        \State Save $LM_{\text{calib}}^{\text{val}}$
        \EndIf
    \EndFor
\end{algorithmic}
\end{algorithm}

\begin{figure}[t]
  \centering
  \includegraphics[width=0.95\linewidth]{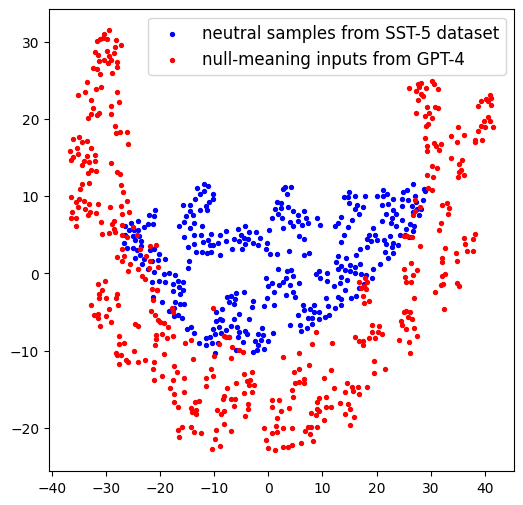}
  \caption{t-SNE visualization of representations for null-meaning inputs generated from GPT-4 (red) compared to neutral samples from SST-5 dataset (blue). We utilize the pre-trained sentiment analysis model \cite{loureiro-etal-2022-timelms} to obtain the embeddings. The different distributions validate that null information is not equivalent to neutral sentiment.}
  \label{fig6}
\end{figure}

\begin{figure*}[t]
  \centering
  \includegraphics[width=1\linewidth]{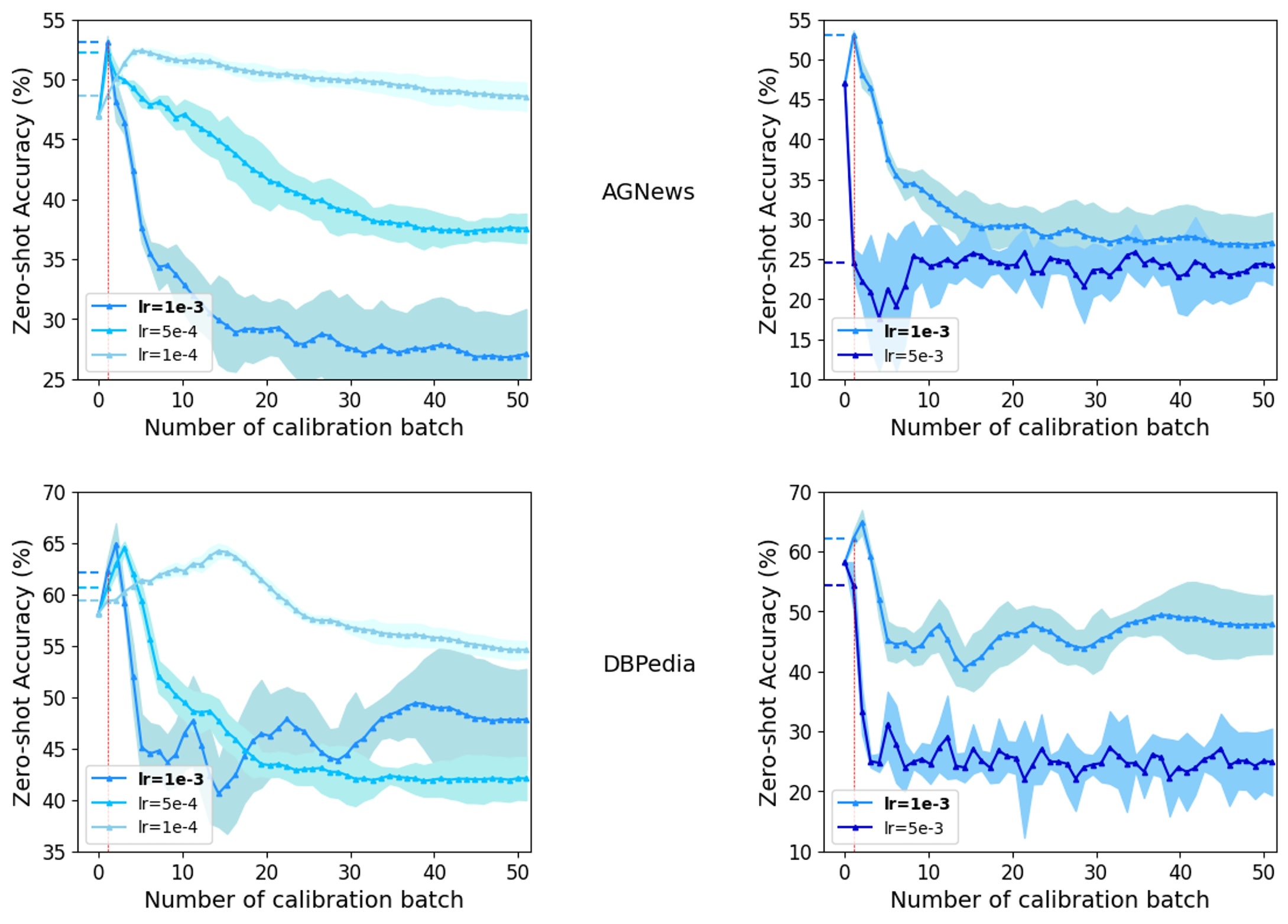}
  \caption{Empirical experiments show the impact of calibration on zero-shot learning performance across \textit{different calibration learning rates $lr_{calib}$}, with a fixed batch size of 32. Only \textbf{\textit{B\textsubscript{LM}}} is updated in calibration. We identify the optimal $lr_{calib}= 1e-3$ across all datasets and illustrate with AGNews dataset (top two figures) and DBPedia dataset (bottom two figures). A smaller learning rate (left figures) consistently yields less performance improvement, considering both peak accuracy and accuracy after the first calibration batch (the intersections of the curves and red vertical line). A larger learning rate (right figures) consistently degrades performance.}
  \label{fig7}
\end{figure*}

\clearpage
\section{Additional Results}
\label{add}

\subsection{Performance Comparison with NSP-BERT, Perplection and  NoisyTune}
\label{compare}
We additionally choose NSP-BERT \cite{sun-etal-2022-nsp} and Perplection \cite{lu-etal-2023-makes} as \textit{in-context learning} comparison baselines and NoisyTune \cite{wu-etal-2022-noisytune} as \textit{prompt-base fine-tuning} comparison baseline. NSP-BERT constructs potential answers using each label word and predict Next Sentence Prediction (NSP) probability between the input and each answer. Perplection proposes perplexity-based selection method for prompt-based zero-shot learning. NoisyTune demonstrates that adding noise to pre-trained LMs benefits fine-tuning on downstream tasks. We re-implement their methods with the same settings as ours for fair comparisons. As shown in Table~\ref{table16} and Table~\ref{table5}, our method achieves superior results in almost all datasets.

Furthermore, our method consistently outperforms NoisyTune, demonstrating that the gains in prompt-based fine-tuning with our method are not solely a result of perturbing LM parameters. This confirms the efficacy of intrinsic bias calibration in enhancing LM performance.

\begin{table}[H]
    \centering
    \small
    \setlength{\tabcolsep}{8pt}
    \begin{tabular}{c c c c}
        \toprule
        \multirow{2.5}{*}{\textbf{}} & 
        \multicolumn{3}{c}{\textbf{Zero-shot in-context learning}}\\
        \\[-2ex]
        \cline{2-4}
        \\[-1.5ex]
         & NSP-BERT & Perplection & IntrCal \\
        \midrule
        AGNews       & $52.4$ & $49.3$ & $\textbf{54.5}$ \\[0.5ex]
        DBPedia      & $58.4$ & $59.6$ & $\textbf{61.8}$ \\[0.5ex]
        TREC         & $\textbf{32.4}$ & $30.8$ & $31.1$ \\[0.5ex]
        Subj         & $60.3$ & $59.9$ & $\textbf{62.7}$ \\[0.5ex]
        SST-5        & $30.2$ & $31.0$ & $\textbf{37.5}$ \\[0.5ex]
        Laptop       & $57.3$ & $58.2$ & $\textbf{59.6}$ \\[0.5ex]
        Restaurant   & $50.4$ & $66.5$ & $\textbf{72.8}$ \\[0.5ex]
        Twitter      & $35.3$ & $31.5$ & $\textbf{51.7}$ \\[0.001ex]
        \midrule
        Average & $47.1$ & $48.4$ & $\textbf{54.0}$ \\[0.001ex]
        \bottomrule
    \end{tabular}
    \caption{Comparison of zero-shot in-context learning performance across NSP-BERT \cite{sun-etal-2022-nsp}, Perplection \cite{lu-etal-2023-makes} and IntrCal (ours).}
    \label{table16}
\end{table}

\begin{table}[t]
    \centering
    \small
    \setlength{\tabcolsep}{2pt}
    \begin{tabular}{c cc c cc}
        \toprule
        \multirow{2.5}{*}{\textbf{}} & 
        \multicolumn{2}{c}{\textbf{Prompt FT no demo}} & &
        \multicolumn{2}{c}{\textbf{Prompt FT with demo}} \\
        \\[-2ex]
        \cline{2-3} \cline{5-6}
        \\[-1.5ex]
        &  NoisyTune & IntrCal & & NoisyTune & IntrCal \\
        \midrule
        AGNews & $89.0_\text{1.8}$ & $\textbf{89.0}_\text{0.8}$
        & & $88.4_\text{1.5}$ & $\textbf{89.3}_\text{0.9}$ \\[0.5ex]
        
        DBPedia & $98.0_\text{0.8}$ & $\textbf{99.0}_\text{0.1}$
        & & $98.6_\text{0.9}$ & $\textbf{98.9}_\text{0.3}$ \\[0.5ex]
        
        TREC & $86.2_\text{4.3}$ & $\textbf{89.3}_\text{4.5}$
        & & $87.2_\text{4.6}$ & $\textbf{89.7}_\text{1.0}$ \\[0.5ex]
        
        Subj & $93.0_\text{1.2}$ & $\textbf{93.2}_\text{1.2}$
        & & $92.9_\text{1.2}$ & $\textbf{94.3}_\text{0.2}$ \\[0.5ex]
        
        SST-5 & $49.4_\text{1.1}$ & $\textbf{49.9}_\text{2.7}$
        & & $47.5_\text{3.5}$ & $\textbf{50.0}_\text{1.7}$ \\[0.5ex]
        
        Laptop & $73.8_\text{3.2}$ & $\textbf{74.9}_\text{2.9}$
        & & $75.5_\text{3.2}$ & $\textbf{78.7}_\text{1.4}$ \\[0.5ex]
        
        Restaurant & $79.9_\text{2.7}$ & $\textbf{82.0}_\text{0.9}$
        & & $78.3_\text{2.6}$ & $\textbf{79.8}_\text{4.5}$ \\[0.5ex]
        
        Twitter & $51.8_\text{5.8}$ & $\textbf{57.0}_\text{4.2}$
        & & $59.0_\text{1.9}$ & $\textbf{59.3}_\text{2.3}$ \\[0.001ex]
        
        \midrule
        Average & $77.6$ & $\textbf{79.3}$ 
        & & $78.4$ & $\textbf{80.0}$ \\
        \bottomrule
    \end{tabular}
    \caption{Comparison between NoisyTune \cite{wu-etal-2022-noisytune} and IntrCal (ours) in prompt-based fine-tuning.}
    \label{table5}
\end{table}

\subsection{Effectiveness on Decoder LMs}
\label{decoder}
We validate the effectiveness of intrinsic bias calibration in enhancing prompt-based learning performance on GPT-2 XL (1.5B) and Llama-2 (7B). The same hyper-parameters from Appendix~\ref{details} and prompt templates from Table~\ref{table7} are used for bias calibration. For GPT-2, we only update the bias parameters during calibration, whereas for Llama-2, we update the entire model since it does not have bias parameters. We conduct zero-shot and two-shot in-context learning experiments across the eight classification datasets, comparing original (Orig.) LM and calibrated (Calib.) LM. The performance comparisons are shown in Table~\ref{gpt-2} (GPT-2) and Table~\ref{llama-2} (Llama-2). Calibrated LMs demonstrate significant performance improvement compared to original pre-trained LMs.

\begin{table}[H]
    \centering
    \small
    \setlength{\tabcolsep}{2.8pt} 
    \begin{tabular}{l c c c c c}
        \toprule
        \multirow{2}{*}{} & \multicolumn{2}{c}{\textbf{Zero-shot}} &  & \multicolumn{2}{c}{\textbf{Two-shot}} \\
        \\[-2ex]
        \cline{2-3} \cline{5-6}
        \\[-1.5ex]
        & Orig. LM & Calib. LM &  & Orig. LM & Calib. LM \\
        \midrule
        AGNews       & $31.5_\text{0.0}$ & $\textbf{41.8}_\text{1.8}$ &  & $74.4_\text{2.6}$ & $\textbf{76.6}_\text{2.5}$ \\[0.5ex]
        DBPedia      & $37.6_\text{0.0}$ & $\textbf{42.1}_\text{1.2}$ &  & $66.8_\text{1.8}$ & $\textbf{70.9}_\text{2.2}$ \\[0.5ex]
        TREC         & $37.0_\text{0.0}$ & $\textbf{40.3}_\text{0.4}$ &  & $42.8_\text{3.1}$ & $\textbf{48.2}_\text{0.6}$ \\[0.5ex]
        Subj         & $50.1_\text{0.0}$ & $\textbf{55.0}_\text{0.1}$ &  & $71.4_\text{3.6}$ & $\textbf{73.0}_\text{2.4}$ \\[0.5ex]
        SST-5        & $33.2_\text{0.0}$ & $\textbf{38.9}_\text{0.4}$ &  & $29.3_\text{0.7}$ & $\textbf{31.1}_\text{0.4}$ \\[0.5ex]
        Laptop       & $39.6_\text{0.0}$ & $\textbf{45.7}_\text{0.4}$ &  & $46.2_\text{4.2}$ & $\textbf{53.1}_\text{2.2}$ \\[0.5ex]
        Restaurant   & $56.6_\text{0.0}$ & $\textbf{63.7}_\text{0.5}$ &  & $66.8_\text{0.9}$ & $\textbf{68.9}_\text{0.6}$ \\[0.5ex]
        Twitter      & $22.7_\text{0.0}$ & $\textbf{38.4}_\text{0.5}$ &  & $29.4_\text{5.4}$ & $\textbf{46.8}_\text{3.2}$ \\[0.001ex]
        \midrule
        Average      & $38.5$            & $\textbf{45.7}$            &  & $53.4$            & $\textbf{58.6}$            \\
    \bottomrule
    \end{tabular}
    \caption{Performance comparison before and after intrinsic bias calibration for \textbf{GPT-2 XL}.}
    \label{gpt-2}
\end{table}

\begin{table}[H]
    \centering
    \small
    \setlength{\tabcolsep}{2.8pt} 
    \begin{tabular}{c c c c c c}
        \toprule
        \multirow{2}{*}{} & \multicolumn{2}{c}{\textbf{Zero-shot}} &  & \multicolumn{2}{c}{\textbf{Two-shot}} \\
        \\[-2ex]
        \cline{2-3} \cline{5-6}
        \\[-1.5ex]
        & Orig. LM & Calib. LM &  & Orig. LM & Calib. LM \\
        \midrule
        AGNews       & $44.1_\text{0.0}$ & $\textbf{50.6}_\text{1.5}$ &  & $80.8_\text{3.4}$ & $\textbf{83.4}_\text{2.4}$ \\[0.5ex]
        DBPedia      & $47.1_\text{0.0}$ & $\textbf{51.2}_\text{0.6}$ &  & $88.5_\text{5.1}$ & $\textbf{93.8}_\text{1.6}$ \\[0.5ex]
        TREC         & $42.0_\text{0.0}$ & $\textbf{44.4}_\text{1.4}$ &  & $51.0_\text{1.2}$ & $\textbf{54.3}_\text{0.5}$ \\[0.5ex]
        Subj         & $49.8_\text{0.0}$ & $\textbf{60.1}_\text{0.3}$ &  & $49.5_\text{6.3}$ & $\textbf{58.4}_\text{2.1}$ \\[0.5ex]
        SST-5        & $29.3_\text{0.0}$ & $\textbf{33.5}_\text{1.2}$ &  & $26.1_\text{4.2}$ & $\textbf{36.4}_\text{3.2}$ \\[0.5ex]
        Laptop       & $48.5_\text{0.0}$ & $\textbf{52.4}_\text{2.3}$ &  & $54.2_\text{3.0}$ & $\textbf{56.1}_\text{1.5}$ \\[0.5ex]
        Restaurant   & $65.4_\text{0.0}$ & $\textbf{70.0}_\text{0.8}$ &  & $59.2_\text{4.1}$ & $\textbf{68.7}_\text{0.8}$ \\[0.5ex]
        Twitter      & $25.5_\text{0.0}$ & $\textbf{42.6}_\text{3.2}$ &  & $27.1_\text{1.4}$ & $\textbf{44.8}_\text{1.9}$ \\[0.001ex]
        \midrule
        Average      & $44.0$            & $\textbf{50.6}$            &  & $54.6$            & $\textbf{62.0}$            \\
    \bottomrule
    \end{tabular}
    \caption{Performance comparison before and after intrinsic bias calibration for \textbf{Llama-2}.}
    \label{llama-2}
\end{table}

In Table~\ref{table15}, we compare the performance of RoBERTa-large (355M) with GPT-2 XL (1.5B) and Llama-2 (7B) in zero-shot learning on classification tasks, using their original pre-trained models. RoBERTa outperforms the other models on more datasets, and achieves better computing efficiency due to its smaller model size. Encoder LMs could be more effective and efficient for classification tasks for several reasons: 
(\romannumeral 1) The bidirectional architecture of encoder LMs enables them to capture task-specific patterns more effectively by attending to both left and right context, compared to the unidirectional nature of decoder LMs. (\romannumeral 2) Classification tasks prioritize accurate label prediction over the generation of diverse and human-like text. Besides, the label spaces in classification are significantly more constrained than the whole vocabulary used in generative applications, which may restrict the effectiveness of decoder LMs \cite{li2023label}. (\romannumeral 3) The relative small size of encoder models facilitates efficiently combining prompting with label-supervised fine-tuning for classification tasks \cite{liu2023pre}, which further enhances performance, as demonstrated in Table~\ref{table2}.

\subsection{Other Experiments}
\label{other_exp}
We briefly summarize the contents of each table and figure below that presents other additional results.

\begin{figure}[t]
  \centering
  \includegraphics[width=1\linewidth]{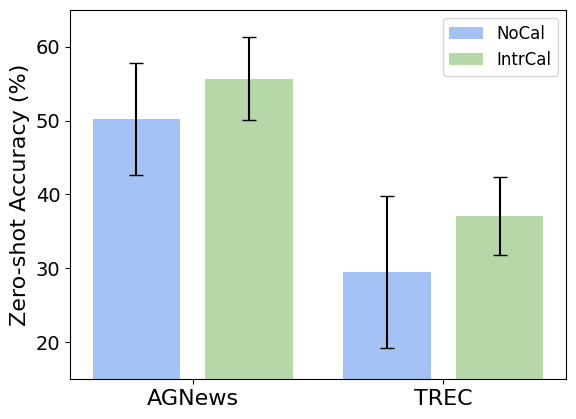}
  \caption{Performance comparison averaged on using five different prompt templates with RoBERTa-large. IntrCal (ours; intrinsic-bias calibrated LM) demonstrates significantly improved accuracy with lower variance compared to NoCal (no calibration).}
  \label{fig8}
\end{figure}

\vspace{10pt}
\noindent Figure~\ref{fig8} contains results for performance using different prompt templates (Table~\ref{table10}).

\vspace{10pt}
\noindent Table~\ref{table8} contains results for performance using RoBERTa-base model.

\vspace{10pt}
\noindent Table~\ref{table9} contains results for performance of $K=\{2, 4, 8\}$ few-shot learning.

\vspace{10pt}
\noindent Table~\ref{table14} contains results for pseudo-perplexity comparisons between updating entire LM and only updating bias parameters in calibration.

\vspace{10pt}
\noindent Table~\ref{table11} contains results for performance comparisons between updating entire LM and only updating bias parameters in calibration.

\vspace{10pt}
\noindent Table~\ref{table17} contains results for performance of sentence-pair datasets.

\vspace{10pt}
\noindent Table~\ref{table12} contains results for variance of probability distribution across labels before and after calibration.
\begin{table}[t]
    \centering
    \small
    \setlength{\tabcolsep}{6pt}
    \begin{tabular}{c c c c}
        \toprule
        & \textbf{RoBERTa-large} & \textbf{GPT-2 XL} & \textbf{Llama-2} \\[0.2ex] 
        & (355M) & (1.5B) & (7B)        \\

        \midrule
        AGNews & $\textbf{47.0}$ & $31.5$ & $44.1$ \\[0.5ex]
        DBPedia & $\textbf{58.2}$ & $37.6$ & $47.1$ \\[0.5ex]
        TREC & $24.0$ & $37.0$ & $\textbf{42.0}$ \\[0.5ex]
        Subj & $\textbf{50.8}$ & $50.1$ & $49.8$ \\[0.5ex]
        SST-5 & $31.5$ & $\textbf{33.2}$ & $29.3$ \\[0.5ex]
        Laptop & $\textbf{54.6}$ & $39.6$ & $48.5$ \\[0.5ex]
        Restaurant & $\textbf{68.6}$ & $56.6$ & $65.4$ \\[0.5ex]
        Twitter & $19.7$ & $22.7$ & $\textbf{25.5}$ \\[0.001ex]
        \midrule
        Average & $\textbf{44.3}$ & $38.5$ & $44.0$ \\[0.001ex]
        \bottomrule
    \end{tabular}
    \caption{Comparison of zero-shot in-context learning performance on classification tasks across RoBERTa-large (355M), GPT-2 XL (1.5B), and Llama-2 (7B).}
    \label{table15}
\end{table}


\begin{table}[t]
    \centering
    \small
    \setlength{\tabcolsep}{14pt}
    \begin{tabular}{c c}
        \toprule
        \textbf{Task} & \textbf{Prompt Templates} \\
        \midrule
        AGNews & \thead{
        \{Sentence\} It is about \texttt{<mask>}.\\[1ex]
        \{Sentence\} This is about \texttt{<mask>}.\\[1ex]
        \{Sentence\} This is on \texttt{<mask>}.\\[1ex]
        \{Sentence\} It pertains to \texttt{<mask>}.\\[1ex]
        \{Sentence\} In relation to \texttt{<mask>}.}\\
        \midrule
        TREC & \thead{
        \{Sentence\} It is about \texttt{<mask>}. \\[1ex]
        \{Sentence\} Concerning \texttt{<mask>}. \\[1ex]
        \{Sentence\} This is about \texttt{<mask>}. \\[1ex]
        \{Sentence\} In relation to \texttt{<mask>}. \\[1ex]
        \{Sentence\} This is on \texttt{<mask>}.} \\    
        \bottomrule
    \end{tabular}
    \caption{The five different prompt templates used in Figure~\ref{fig8}.}
    \label{table10}
\end{table}

\begin{table*}[t]
    \centering
    \small
    \setlength{\tabcolsep}{3pt}
    \begin{tabular}{c cccc cccc cccc ccc}
        \toprule
        \multirow{2.5}{*}{\textbf{}} & 
        \multicolumn{3}{c}{\textbf{In-context lrn no demo}} & & 
        \multicolumn{3}{c}{\textbf{In-context lrn with demo}} & &
        \multicolumn{3}{c}{\textbf{Prompt FT no demo}} & &
        \multicolumn{3}{c}{\textbf{Prompt FT with demo}} \\
        \\[-2ex]
        \cline{2-4} \cline{6-8} \cline{10-12} \cline{14-16}
        \\[-1.5ex]
        & NoCal & OutCal & IntrCal & & 
        NoCal & OutCal & IntrCal & &
        NoCal & OutCal & IntrCal & &
        NoCal & OutCal & IntrCal \\
        \midrule
        AGNews & $37.8_\text{0.0}$ & $36.2_\text{4.6}$ & $\textbf{49.0}_\text{0.9}$ & & $68.4_\text{0.4}$ & $69.7_\text{4.3}$ & $\textbf{73.7}_\text{0.3}$ 
        & & $88.2_\text{0.3}$ & $87.8_\text{0.6}$ & $\textbf{88.9}_\text{1.0}$ 
        & & $86.7_\text{0.1}$ & $74.2_\text{4.1}$ & $\textbf{87.2}_\text{0.1}$ \\[0.5ex]
        
        DBPedia & $\textbf{57.2}_\text{0.0}$ & $50.5_\text{7.1}$ & $54.9_\text{0.1}$ 
        & & $56.5_\text{3.4}$ & $78.7_\text{4.4}$ & $\textbf{83.9}_\text{0.4}$
        & & $95.2_\text{2.1}$ & $93.5_\text{5.0}$ & $\textbf{99.0}_\text{0.4}$ 
        & & $97.8_\text{0.9}$ & $96.7_\text{0.8}$ & $\textbf{98.6}_\text{0.1}$ \\[0.5ex]
        
        TREC & $28.2_\text{0.0}$ & $25.4_\text{4.4}$ & $\textbf{30.2}_\text{0.1}$ 
        & & $41.2_\text{0.3}$ & $39.9_\text{3.8}$ & $\textbf{42.5}_\text{1.0}$
        & & $82.5_\text{10.9}$ & $70.3_\text{2.3}$ & $\textbf{86.4}_\text{6.5}$
        & & $85.7_\text{1.8}$ & $80.6_\text{5.0}$ & $\textbf{91.2}_\text{0.6}$ \\[0.5ex]
        
        Subj & $53.6_\text{0.0}$ & $63.6_\text{1.9}$ & $\textbf{66.4}_\text{1.8}$ 
        & & $50.8_\text{0.2}$ & $67.0_\text{1.7}$ & $\textbf{69.6}_\text{0.4}$ 
        & & $\textbf{92.5}_\text{1.3}$ & $91.1_\text{0.4}$ & $91.9_\text{1.7}$ 
        & & $90.4_\text{2.1}$ & $92.0_\text{0.2}$ & $\textbf{92.3}_\text{0.1}$ \\[0.5ex]
        
        SST-5 & $31.9_\text{0.0}$ & $30.8_\text{3.4}$ & $\textbf{32.2}_\text{0.2}$ 
        & & $25.3_\text{4.3}$ & $28.6_\text{3.4}$ & $\textbf{29.8}_\text{1.7}$ 
        & & $45.9_\text{3.3}$ & $42.9_\text{2.3}$ & $\textbf{48.1}_\text{1.8}$ 
        & & $44.3_\text{5.2}$ & $40.7_\text{2.5}$ & $\textbf{45.8}_\text{2.6}$ \\[0.5ex]
        
        Laptop & $56.1_\text{0.0}$ & $56.7_\text{3.8}$ & $\textbf{60.0}_\text{0.1}$  
        & & $49.2_\text{0.9}$ & $61.5_\text{2.8}$ & $\textbf{64.0}_\text{0.6}$ 
        & & $75.8_\text{3.4}$ & $73.0_\text{1.3}$ & $\textbf{76.3}_\text{1.8}$ 
        & & $74.8_\text{0.1}$ & $76.0_\text{0.6}$ & $\textbf{76.3}_\text{0.5}$ \\[0.5ex]
        
        Restaurant & $69.8_\text{0.0}$ & $\textbf{72.0}_\text{2.9}$ & $69.5_\text{0.5}$ 
        & & $67.6_\text{0.7}$ & $70.5_\text{2.4}$ & $\textbf{73.2}_\text{0.7}$
        & & $75.5_\text{6.6}$ & $\textbf{77.3}_\text{3.4}$ & $77.2_\text{1.1}$ 
        & & $74.8_\text{3.3}$ & $75.2_\text{0.7}$ & $\textbf{76.1}_\text{3.9}$ \\[0.5ex]
        
        Twitter & $22.0_\text{0.0}$ & $48.6_\text{5.1}$ & $\textbf{52.3}_\text{0.6}$  
        & & $17.6_\text{0.4}$ & $41.8_\text{5.4}$ & $\textbf{48.4}_\text{0.5}$ 
        & & $54.5_\text{1.1}$ & $47.7_\text{3.8}$ & $\textbf{57.9}_\text{1.3}$  
        & & $50.6_\text{4.6}$ & $51.8_\text{2.1}$ & $\textbf{56.0}_\text{4.9}$ \\[0.001ex]
        
        \midrule
        Average & $44.6$ & $48.0$ & $\textbf{51.8}$
        & & $47.1$ & $57.2$ & $\textbf{60.6}$ 
        & & $76.3$ & $73.0$ & $\textbf{78.2}$ 
        & & $75.6$ & $73.4$ & $\textbf{77.9}$ \\
        \bottomrule
    \end{tabular}
    \caption{Result comparisons among NoCal (LM-BFF \citealp{gao-etal-2021-making}; no calibration), OutCal (output calibration) and IntrCal (ours; intrinsic-bias calibrated LM) using \underline{RoBERTa-base}. We report the mean and standard deviation of performance in 8 classification datasets with 4 prompt-based learning methods.}
    \label{table8}
\end{table*}

\begin{table*}[t]
\centering
\small
\setlength{\tabcolsep}{12pt}
\begin{tabular}{cc cc cc cc}
\toprule

\multicolumn{2}{c}{} &
\multicolumn{2}{c}{\textbf{In-context lrn with demo}} & 
\multicolumn{2}{c}{\textbf{Prompt FT no demo}} & 
\multicolumn{2}{c}{\textbf{Prompt FT with demo}} \\

\cmidrule(lr){3-4} \cmidrule(lr){5-6} \cmidrule(lr){7-8}

\multicolumn{2}{c}{} &
\multicolumn{1}{c}{\hspace{1em}NoCal} & \multicolumn{1}{c}{IntrCal} & 
\multicolumn{1}{c}{\hspace{1em}NoCal} & \multicolumn{1}{c}{IntrCal} & 
\multicolumn{1}{c}{\hspace{1em}NoCal} & \multicolumn{1}{c}{IntrCal} \\

\midrule

\multirow{4.5}{*}{2-shot} 
& AGNews 
& \hspace{1em} $70.4_\text{6.7}$ & $\textbf{76.3}_\text{3.6}$ 
& \hspace{1em} $76.4_\text{5.4}$ & $\textbf{80.2}_\text{8.0}$ 
& \hspace{1em} $78.2_\text{1.3}$ & $\textbf{83.2}_\text{1.1}$ \\[0.5ex]

& DBPedia 
& \hspace{1em} $92.9_\text{0.9}$ & $\textbf{94.0}_\text{1.0}$ 
& \hspace{1em} $97.0_\text{1.6}$ & $\textbf{98.4}_\text{0.9}$ 
& \hspace{1em} $97.4_\text{1.0}$ & $\textbf{97.8}_\text{1.1}$ \\[0.5ex]

& TREC 
& \hspace{1em} $49.8_\text{4.2}$ & $\textbf{50.5}_\text{4.0}$ 
& \hspace{1em} $49.1_\text{22.6}$ & $\textbf{60.3}_\text{9.6}$ 
& \hspace{1em} $65.2_\text{9.3}$ & $\textbf{66.1}_\text{9.3}$ \\[0.5ex]

& Subj
& \hspace{1em} $49.4_\text{1.1}$ & $\textbf{56.2}_\text{3.9}$ 
& \hspace{1em} $66.4_\text{5.4}$ & $\textbf{82.2}_\text{5.9}$ 
& \hspace{1em} $72.3_\text{13.9}$ & $\textbf{81.5}_\text{13.2}$ \\

\midrule

\multirow{4.5}{*}{4-shot} 
& AGNews 
& \hspace{1em} $75.7_\text{3.9}$ & $\textbf{80.3}_\text{1.7}$ 
& \hspace{1em} $85.4_\text{2.7}$ & $\textbf{87.3}_\text{1.3}$  
& \hspace{1em} $76.7_\text{13.1}$ & $\textbf{85.9}_\text{1.9}$ \\[0.5ex]

& DBPedia 
& \hspace{1em} $93.0_\text{0.4}$ & $\textbf{93.9}_\text{0.4}$ 
& \hspace{1em} $97.2_\text{0.8}$ & $\textbf{97.9}_\text{1.1}$  
& \hspace{1em} $96.4_\text{1.5}$ & $\textbf{98.6}_\text{0.6}$ \\[0.5ex]

& TREC 
& \hspace{1em} $51.9_\text{2.6}$ & $\textbf{53.2}_\text{2.5}$ 
& \hspace{1em} $64.5_\text{7.1}$ & $\textbf{67.6}_\text{6.7}$ 
& \hspace{1em} $73.6_\text{8.5}$ & $\textbf{78.2}_\text{9.7}$ \\[0.5ex]

& Subj
& \hspace{1em} $48.8_\text{2.2}$ & $\textbf{59.4}_\text{3.1}$ 
& \hspace{1em} $81.4_\text{3.9}$ & $\textbf{88.5}_\text{3.2}$ 
& \hspace{1em} $78.9_\text{9.3}$ & $\textbf{83.6}_\text{7.8}$ \\[0.001ex]

\midrule

\multirow{4.5}{*}{8-shot} 
& AGNews 
& \hspace{1em} $79.6_\text{1.0}$ & $\textbf{82.4}_\text{1.6}$ 
& \hspace{1em} $86.9_\text{1.9}$ & $\textbf{88.1}_\text{0.4}$  
& \hspace{1em} $85.5_\text{1.7}$ & $\textbf{88.0}_\text{1.4}$ \\[0.5ex]

& DBPedia 
& \hspace{1em} $92.9_\text{0.8}$ & $\textbf{94.2}_\text{0.2}$ 
& \hspace{1em} $97.3_\text{1.2}$ & $\textbf{98.8}_\text{0.5}$   
& \hspace{1em} $98.2_\text{0.8}$ & $\textbf{98.6}_\text{0.2}$ \\[0.5ex]

& TREC 
& \hspace{1em} $47.9_\text{2.2}$ & $\textbf{48.7}_\text{2.0}$ 
& \hspace{1em} $71.6_\text{4.9}$ & $\textbf{72.2}_\text{5.1}$ 
& \hspace{1em} $75.4_\text{6.2}$ & $\textbf{81.7}_\text{5.6}$ \\[0.5ex]

& Subj
& \hspace{1em} $48.4_\text{1.0}$ & $\textbf{60.5}_\text{4.8}$ 
& \hspace{1em} $91.9_\text{1.3}$ & $\textbf{92.7}_\text{0.8}$ 
& \hspace{1em} $88.9_\text{5.3}$ & $\textbf{92.1}_\text{2.2}$ \\[0.001ex]

\bottomrule
\end{tabular}
\caption{Few-shot learning with different number of training samples ($K=\{2, 4, 8\}$) using RoBERTa-large. IntrCal (ours; intrinsic-bias calibrated LM) consistently outperforms NoCal (no calibration).}
\label{table9}
\end{table*}

\begin{table*}[t]
\centering
\small
\setlength{\tabcolsep}{6.5pt}
\begin{tabular}{l rr r rr r rr}     
\toprule
\multirow{2.5}{*}{\textbf{Model}} & \multicolumn{8}{c}{\textbf{Datasets}} \\

\\[-2ex]
\cline{2-9}
\\[-1.5ex]
\textbf{} & \multicolumn{2}{c}{\textbf{WikiText-2}} & & \multicolumn{2}{c}{\textbf{WikiText-103}} & & \multicolumn{2}{c}{\textbf{LAMBADA}}\\
\midrule
Original RoBERTa  & \multicolumn{2}{c}{$6.189$} & &  \multicolumn{2}{c}{$7.008$} & &  \multicolumn{2}{c}{$24.52$} \\[0.5ex]

\\[-2.5ex]
\cdashline{2-9}[1pt/1pt]
\\[-1.5ex]

\, + \textsc{calibration} & \textbf{\textit{W\textsubscript{LM}}} + \textbf{\textit{B\textsubscript{LM}}} & \textbf{\textit{B\textsubscript{LM}}} & &
\textbf{\textit{W\textsubscript{LM}}} + \textbf{\textit{B\textsubscript{LM}}} & \textbf{\textit{B\textsubscript{LM}}} & &
\textbf{\textit{W\textsubscript{LM}}} + \textbf{\textit{B\textsubscript{LM}}} & \textbf{\textit{B\textsubscript{LM}}}\\[0.5ex]

\\[-2.5ex]
\cdashline{2-9}[1pt/1pt]
\\[-1.5ex]

\, \, \, \text{for\_AGNews} &   \ua{0.105} $6.294$ & \ua{0.017} $6.206$ & & \ua{0.059} $7.067$ & \ua{0.029} $7.037$ & & \ua{0.58} $25.10$ & \ua{0.02} $24.54$ \\[0.5ex]
\, \, \, \text{for\_DBPedia} &  \ua{0.101} $6.290$ & \ua{0.008} $6.197$ & & \ua{0.092} $7.100$ & \ua{0.002} $7.010$ & & \ua{0.76} $25.28$ & \da{0.22} $24.30$ \\[0.5ex]
\, \, \, \text{for\_TREC} &     \ua{0.049} $6.238$ & \da{0.027} $6.162$ & & \ua{0.040} $7.048$ & \da{0.042} $6.966$ & & \ua{0.57} $25.09$ & \da{0.27} $24.25$ \\[0.5ex]
\, \, \, \text{for\_Subj} &     \ua{0.081} $6.270$ & \da{0.021} $6.168$ & & \ua{0.116} $7.124$ & \da{0.030} $6.978$ & & \ua{0.70} $25.22$ & \ua{0.08} $24.60$ \\[0.5ex]
\, \, \, \text{for\_SST-5} &    \da{0.018} $6.171$ & \da{0.031} $6.158$ & & \ua{0.143} $7.151$ & \da{0.039} $6.969$ & & \ua{0.65} $25.17$ & \da{0.18} $24.34$ \\[0.5ex]
\, \, \, \text{for\_Laptop} &   \ua{0.133} $6.322$ & \ua{0.011} $6.200$ & & \ua{0.075} $7.083$ & \ua{0.002} $7.010$ & & \ua{0.56} $25.08$ & \da{0.01} $24.51$ \\[0.5ex]
\, \, \, \text{for\_Restaurant}&\ua{0.102} $6.291$ & \ua{0.055} $6.244$ & & \ua{0.071} $7.079$ & \ua{0.074} $7.082$ & & \ua{0.64} $25.16$ & \ua{0.13} $24.65$ \\[0.5ex]
\, \, \, \text{for\_Twitter} &  \ua{0.204} $6.393$ & \da{0.029} $6.160$ & & \ua{0.096} $7.104$ & \da{0.037} $6.971$ & & \ua{0.39} $24.91$ & \ua{0.05} $24.57$ \\[0.001ex]

\bottomrule
\end{tabular}
\caption{Pseudo-perplexities of original RoBERTa and task-specific calibrated RoBERTa on WikiText-2, WikiText-103 and LAMBADA. We use 2000 test samples of each dataset. An increase in values (highlighted in red) indicates a reduction in language modeling abilities after calibration. \textbf{\textit{W\textsubscript{LM}}} + \textbf{\textit{B\textsubscript{LM}}} updates entire LM in calibration while \textbf{\textit{B\textsubscript{LM}}} only updates bias parameters.}
\label{table14}
\end{table*}

\clearpage
\begin{table}[t]
    \centering
    \small
    \setlength{\tabcolsep}{2.5pt}
    \begin{tabular}{c cc c cc}
        \toprule
        \multirow{2.5}{*}{} & 
        \multicolumn{2}{c}{\textbf{ICL with demo}} & &
        \multicolumn{2}{c}{\textbf{Prompt FT with demo}} \\
        \\[-2ex]
        \cline{2-3} \cline{5-6}
        \\[-1.5ex]
        &  \textbf{\textit{W\textsubscript{LM}}} + \textbf{\textit{B\textsubscript{LM}}} & \textbf{\textit{B\textsubscript{LM}}} & &
        \textbf{\textit{W\textsubscript{LM}}} + \textbf{\textit{B\textsubscript{LM}}} & \textbf{\textit{B\textsubscript{LM}}} \\
        \midrule
        AGNews & $82.0_\text{0.8}$ & $\textbf{82.4}_\text{0.9}$
        & & $\textbf{89.3}_\text{0.6}$ & $89.3_\text{0.9}$ \\[0.5ex]
        
        DBPedia & $\textbf{95.1}_\text{0.7}$ & $94.8_\text{0.7}$ 
        & & $\textbf{99.0}_\text{0.1}$ & $98.9_\text{0.3}$ \\[0.5ex]
        
        TREC & $\textbf{49.1}_\text{2.6}$ & $48.6_\text{2.2}$ 
        & & $88.9_\text{2.3}$ & $\textbf{89.7}_\text{1.0}$ \\[0.5ex]
        
        Subj & $\textbf{65.6}_\text{0.4}$ & $63.5_\text{2.3}$ 
        & & $93.9_\text{1.6}$ & $\textbf{94.3}_\text{0.2}$ \\[0.5ex]
        
        SST-5 & $\textbf{37.1}_\text{1.0}$ & $36.6_\text{1.0}$ 
        & & $\textbf{51.3}_\text{1.7}$ & $50.0_\text{1.7}$ \\[0.5ex]
        
        Laptop & $65.8_\text{0.3}$ & $\textbf{67.4}_\text{1.7}$ 
        & & $77.7_\text{0.8}$ & $\textbf{78.7}_\text{1.4}$ \\[0.5ex]
        
        Restaurant & $72.7_\text{1.2}$ & $\textbf{74.0}_\text{1.0}$ 
        & & $\textbf{81.4}_\text{3.4}$ & $79.8_\text{4.5}$ \\[0.5ex]
        
        Twitter & $45.8_\text{2.7}$ & $\textbf{49.4}_\text{2.7}$ 
        & & $\textbf{60.4}_\text{1.7}$ & $59.3_\text{2.3}$ \\[0.001ex]
    
        \midrule 
        Average & $64.2$ & $\textbf{64.6}$ 
        & & $\textbf{80.2}$ & $80.0$ \\
        \bottomrule
    \end{tabular}
    \caption{Performance comparisons between differently calibrated LMs using RoBERTa-large. ICL stands for in-context learning. \textbf{\textit{W\textsubscript{LM}}} + \textbf{\textit{B\textsubscript{LM}}} updates entire LM in calibration while \textbf{\textit{B\textsubscript{LM}}} only updates bias parameters. This table (prompt-based learning \textit{with} demonstrations) is the supplement to \S~\ref{update} Table~\ref{table3} (prompt-based learning \textit{without} demonstrations).}
    \label{table11}
\end{table}

\begin{table}[t]
    \centering
    \small
    \setlength{\tabcolsep}{3pt}
    \begin{tabular}{c cc c cc}
        \toprule
        \multirow{2.5}{*}{\textbf{}} & 
        \multicolumn{2}{c}{\textbf{In-context lrn no demo}} & &
        \multicolumn{2}{c}{\textbf{Prompt FT no demo}} \\
        \\[-2ex]
        \cline{2-3} \cline{5-6}
        \\[-1.5ex]
        & NoCal & IntrCal & & NoCal & IntrCal\\
        \midrule
        MNLI & $32.7_\text{0.0}$ & $\textbf{37.7}_\text{0.7}$
        & & $67.9_\text{2.1}$ & $\textbf{68.6}_\text{1.9}$\\[0.5ex]
        
        SNLI & $33.6_\text{0.0}$  & $\textbf{36.7}_\text{0.9}$ 
        & & $77.4_\text{2.8}$ & $\textbf{78.5}_\text{2.3}$\\[0.5ex]
        
        MRPC & $51.1_\text{0.0}$ & $\textbf{53.6}_\text{0.2}$ 
        & & $73.6_\text{4.3}$ & $\textbf{74.9}_\text{1.4}$\\[0.5ex]
        
        QQP & $50.8_\text{0.0}$ & $\textbf{54.6}_\text{0.2}$
        & & $65.2_\text{3.5}$ & $\textbf{66.2}_\text{3.3}$\\
        \bottomrule
    \end{tabular}
    \caption{Benchmark on sentence-pair datasets, MNLI \cite{williams-etal-2018-broad}, SNLI \cite{bowman-etal-2015-large}, MRPC \cite{dolan-brockett-2005-automatically}, QQP \cite{wang-etal-2018-glue}. NoCal denotes no-calibration (baseline) and IntrCal denotes our method. Our method demonstrates effectiveness on sentence-pair datasets. The overall low performance of in-context learning can be attributed to two main factors: (1) RoBERTa's inherent limited capabilities when using in-context learning for the more difficult tasks, which is significantly improved with prompt-based fine-tuning. (2) The misalignment between these sentence-pair datasets and the use of single-sentence null inputs for calibration, which could impact the effectiveness of calibration.}
    \label{table17}
\end{table}

\begin{table}[t]
    \centering
    \small
    \setlength{\tabcolsep}{4pt}
    \begin{tabular}{c c c c c c}
        \toprule
         & AGNews & DBPedia & TREC & Subj & SST-5 \\
        \midrule
        Orig. LM     & 0.033 & 0.130 & 0.025 & 0.195 & 0.011 \\[1ex]
        Calib. LM & 0.022 & 0.025 & 0.011 & 0.112 & 0.011 \\[0.001ex]
        \bottomrule
    \end{tabular}
    \caption{We calculate the \textbf{variance} of probability distribution across labels conditioned on null-meaning inputs, i.e., $Var\left(\bar{P}_{\mathcal{X}_{\text{null}}}(\mathcal{Y})\right)$, before and after calibration. A smaller variance indicates that a distribution is closer to uniform distribution. Orig. LM denotes original LM, and Calib. LM denotes the LM after \textit{One-batch Calibration} (\S~\ref{early stop}). The reduced variance after bias calibration demonstrates that our method promotes LM towards a more equitable starting point.}
    \label{table12}
\end{table}

\end{document}